\documentclass[pdflatex,sn-mathphys-num]{sn-jnl}

\usepackage{graphicx}%
\usepackage{multirow}%
\usepackage{amsmath,amssymb,amsfonts}%
\usepackage{amsthm}%
\usepackage{mathrsfs}%
\usepackage[title]{appendix}%
\usepackage{xcolor}%
\usepackage[table]{xcolor}
\usepackage{colortbl}
\usepackage{textcomp}%
\usepackage{manyfoot}%
\usepackage{booktabs}%
\usepackage{enumitem}
\usepackage{algorithm}%
\usepackage{algorithmicx}%
\usepackage{algpseudocode}%
\usepackage{listings}%
\usepackage{todonotes}
\usepackage{tcolorbox}
\usepackage{makecell}
\usepackage{chngcntr}   
\usepackage{etoolbox}    
\counterwithout{table}{section}  
\newif\iflightpdf
\lightpdftrue   

\theoremstyle{thmstyleone}%
\theoremstyle{thmstyletwo}%

\theoremstyle{thmstylethree}%

\begin{document}

\title[Article Title]{Understanding Mechanistic Role of Structural and Functional Connectivity in Tau Propagation Through Multi-Layer Modeling}


\author[1]{\fnm{Tingting} \sur{Dan}}\email{tdan@email.unc.edu}
\equalcont{These authors contributed equally to this work.}

\author[2]{\fnm{Xinwei} \sur{Huang}}\email{huanggenetics@tongji.edu.cn}
\equalcont{These authors contributed equally to this work.}

\author[1,3]{\fnm{Jiaqi} \sur{Ding}}\email{jiaqid@cs.unc.edu}

\author[2]{\fnm{Yinggang} \sur{Zheng}}\email{ygzheng@tongji.edu.cn}

\author*[1,3,4]{\fnm{Guorong} \sur{Wu}}\email{grwu@med.unc.edu}

\affil[1]{\orgdiv{Department of Psychiatry}, \orgname{University of North Carolina at Chapel Hill}, \orgaddress{\street{333 S. Columbia Street}, \city{Chapel Hill}, \postcode{27599}, \state{NC}, \country{United States of America}}}

\affil[2]{\orgdiv{School of Medicine}, \orgname{Tongji University}, \orgaddress{\city{Shanghai}, \postcode{200434}, \country{China}}}

\affil[3]{\orgdiv{Department of Computer Science}, \orgname{University of North Carolina at Chapel Hill}, \orgaddress{\street{201 S. Columbia St}, \city{Chapel Hill}, \postcode{27599}, \state{NC}, \country{United States of America}}}

\affil[4]{\orgdiv{Department of Statistics and Operations Research (STOR), Carolina Institute for Developmental Disabilities, and the UNC NeuroScience Center}, \orgname{University of North Carolina at Chapel Hill}, \orgaddress{\city{Chapel Hill}, \postcode{27599}, \state{NC}, \country{United States of America}}}

\abstract{
Emerging neuroimaging evidence shows that pathological tau proteins build up along specific brain networks, suggesting that large-scale network architecture plays a key role in the progression of Alzheimer's disease (AD). However, the extent to which structural connectivity (SC) and functional connectivity (FC) interact to influence the propagation of tau aggregates is largely unexplored. Leveraging an unprecedented volume of longitudinal neuroimaging data, we investigate the roles of structural and functional connectivity in tau propagation using a multi-layer graph diffusion model that explicitly incorporates SC–FC interactions. In addition to the evidence that tau propagation is shaped by connectome architectures, our model reveals a regionally asymmetric contribution of SC and FC. Specifically, FC predominantly drives tau spread in subcortical areas, the insula, frontal and temporal cortices, whereas SC plays a more significant role in occipital, parietal, and limbic regions. 
However, the relative dominance of SC versus FC in driving tau propagation shifts throughout the course of disease progression, with FC generally playing a predominant role in the early stages of AD, followed by a shift toward SC as the primary driver in later stages.
Furthermore, the spatial patterns of SC- and FC-dominant regions involved in tau propagation show strong correspondence with the regional expression of several AD-associated genes implicated in inflammation, apoptosis, and lysosomal function, including CHUK (IKK-$\alpha$), TMEM106B, MCL1, NOTCH1, and TH. 
In parallel, we have discovered that other non-modifiable risk factors (such as APOE genotype and biological sex) and biological mechanisms (such as amyloid deposition) can selectively alter the landscape of tau propagation by shifting dominant routes from anatomical to functional pathways (or vice versa) in a regionally specific manner. Our discoveries have been validated in an independent AD cohort, yielding consistent findings. 

}





\maketitle

\section{Introduction}\label{sec1}
Alzheimer's disease (AD) is a progressive neurodegenerative disorder marked by the pathological accumulation and propagation of tau proteins \cite{braak1991neuropathological}. Tau pathology, initially localized in the entorhinal cortex, gradually spreads to connected brain regions in a pattern that correlates with cognitive decline and disease severity. Mounting evidence supports the hypothesis that this spread occurs in a prion-like manner, whereby misfolded tau seeds propagate trans-synaptically through neural networks \cite{palop2016network}. This insight has shifted the focus from region-specific atrophy to network-based degeneration, assuming the brain’s large-scale connectivity architecture as a key determinant of pathological progression.

In this regard, a growing body of research has developed connectome-based diffusion models to simulate and predict the spatial and temporal dynamics of tau propagation across the brain \cite{raj2012network,vogel2020spread,dan2024tauflownet}. These models typically leverage information from structural connectivity (SC), derived from diffusion‑weighted imaging  (DWI), or functional connectivity (FC), measured via resting-state functional magnetic resonance imaging (fMRI), to model tau spread as a network-driven diffusion process. By modeling tau dynamics within the topology of the brain connectome, these frameworks offer a mechanistic perspective on how pathological burden evolves across regions. Given the connectome-constrained assumption on tau propagation, such models have shown potential in forecasting future tau accumulation \cite{dan2024tauflownet,raj2012network}, identifying vulnerable brain circuits \cite{dan2024tauflownet}, and stratifying individuals based on progression risk \cite{vogel2021four}, thereby informing early diagnosis and therapeutic targeting.

While recent computational approaches have achieved notable success in forecasting future pathological burdens, they generally do not model pathology propagation by jointly considering both structural connectivity (SC) and functional connectivity (FC), let alone their interactions. As a result, these methods often fail to reflect the true complexity of disease spread, neglecting the synergistic and potentially nonlinear dynamics linking anatomical connections and neural activity. Given the growing body of evidence that both SC and FC influence the dissemination of pathological processes, elucidating the mechanistic role of their interplay is essential for understanding the progression of neurodegenerative disorders.


To that end, we investigate the joint effect of SC and FC on the propagation of tau aggregates using a multi-layer network diffusion model \cite{huang2025multilayer}, which is trained to predict the trajectory of tau evolution underlying SC-FC topology.
Our framework is grounded in the physical principles of graph heat diffusion, where tau accumulation is modeled as a dynamic process governed by the topology of the underlying connectome. This principled foundation not only enhances interpretability but also offers a biologically plausible mechanism for simulating how pathology propagates through complex multi-layer brain networks.
By disentangling the contributions of structural and functional connectivity to tau accumulation, our computational model enables us to address the following key questions: (1) What is the interplay between SC and FC on the change of tau aggregates at each brain region? (2) What is the role of non-modifiable factors, such as AD-relevant risk genes and biological sex, in the mechanism of SC-FC on tau propagation? (3) Does SC/FC contribution to tau propagation vary during disease progression? and (4) Does the accumulation of A$\beta$ plaque modify the joint effect of SC and FC on tau propagation?


We have applied our computational model on longitudinal tau-PET images in the ADNI dataset \cite{mueller2005ways} and the OASIS \cite{lamontagne2020oasis4} datasets separately. The most prominent and reproducible findings are as follows:
\begin{itemize}
    \item Tau propagation is shaped by both SC and FC. However, the dynamic process of tau aggregation exhibits region-specific selectivity, with some regions predominantly influenced by SC and others by FC. 
    \item The SC-FC selective mechanism evolves throughout the course of disease progression. FC generally predominates in the early stages, while SC becomes more influential as the disease advances. 
    \item At the population level, the spatial distribution of SC-FC selectivity closely aligns with the expression profiles of key AD risk genes, such as CHUK (IKK-$\alpha$), TMEM106B, MCL1, NOTCH1, and TH, known to regulate inflammation, apoptosis, and lysosomal pathways.  
    \item Both non-modifiable factors and the presence of other AD pathology can alter the spatial pattern of SC-FC selective mechanism for tau propagation. For example, APOE genotype alters the landscape of tau propagation by shifting dominant routes from anatomical to functional pathways (or vice versa) in a regionally specific manner. Excessive A$\beta$ deposition promotes a shift in tau propagation dynamics, making SC the predominant factor in the frontal cortex. 
    \item Biological sex does not substantially alter SC–FC dominance, except in the occipital cortex where males show a slight shift from SC- to FC-dominant propagation. 
\end{itemize}

Furthermore, we investigate the causal hypotheses of whether SC influences tau propagation through FC (or vice versa), which allows us to effectively inform therapeutic targets. For example, if the effect of FC on tau propagation is mediated by SC, interventions aimed at preserving axonal integrity or white matter health may prove more effective.
Together, our findings provide novel insights into the mechanistic pathways that govern tau propagation, with potential applications in evaluating risk, tracking progression, and informing the design of targeted interventions.

\section{Results} 
\textbf{Prologue.} \textit{First}, an overview of subject demographics and imaging data characteristics is provided in Section \ref{sec_data}. \textit{Second}, we present the evidence supporting that the evolution of tau aggregation throughout the human brain is shaped by both structural and functional connectome architecture. Additionally, our model reveals that tau propagation is not a single-phase phenomenon, but rather a dynamic, multi-stage process over time. Together, the supporting evidence presented in Section \ref{sec_supportI} and \ref{sec_supportII} sets the stage for investigating the mechanistic role of SC and FC in tau propagation, for conducting stratified analysis by age, APOE4 status, biological sex, and A$\beta$ burden, and for underlying genetic mechanisms in Section \ref{sec_main}. \textit{Third}, we identify the potential causal pathways between the interplay between SC-derived tau aggregation, FC-derived tau aggregation, and cognitive decline, in Section \ref{sec_medi}, 




\subsection{Summary statistics of tau evolution}
\label{sec_data}
\begin{figure}[h]
\centering
\includegraphics[width=1.02\textwidth]{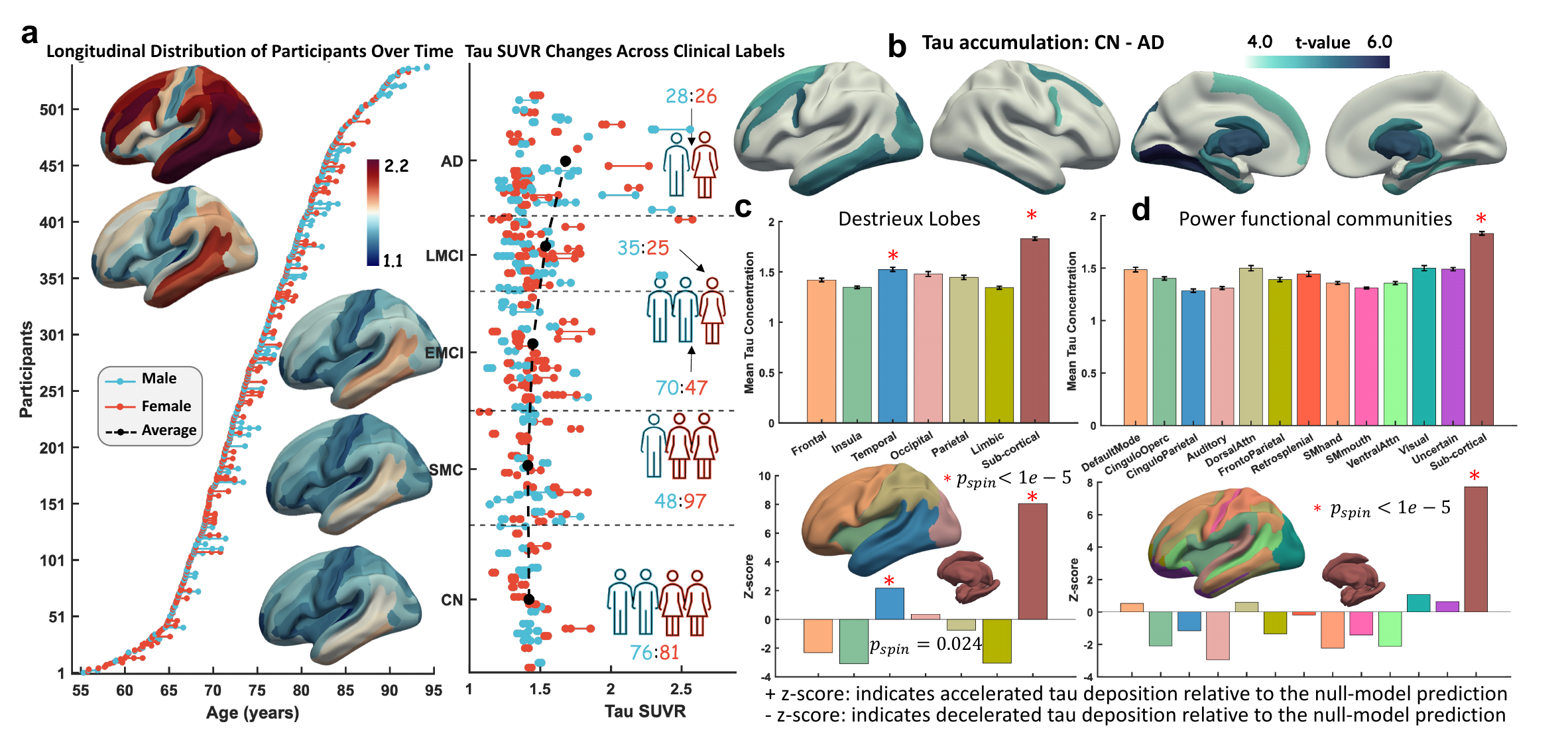}
\caption[]{
\textbf{Typical trajectory of tau evolution in aging brains.}  
\textbf{(a)} Left: Scatter of participant ages at longitudinal tau PET scan, with lines linking multiple scans per individual. The cortical mappings represent the average tau SUVR for each clinical cohort. Right: Regional tau SUVR values grouped by clinical diagnosis (CN, SMC, EMCI, LMCI, AD), with black dots marking group means, illustrating the progressive increase in tau burden across disease stages.
\textbf{(b)} Surface map of regional SUVR differences (CN vs.\ AD) by a linear mixed‐effects model  (random intercept per subject; sex covariate; Bonferroni‐corrected $p<5\times10^{-5}$, two‐sided). Higher positive $t$‐values (darker blue) indicate stronger tau propagation.  
\textbf{(c,d)} Community‐level aggregation of tau progression. Regions were assigned to anatomical lobes based on the Destrieux atlas \cite{destrieux2010automatic} (\textbf{c}) and to functional modules by Power \emph{et al.} \cite{power2011functional} (\textbf{d}). For each community, mean $t$‐values were converted to $z$‐scores via 1,000 spin‐permutation tests \footnotemark[1] to correct for spatial autocorrelation \cite{alexander2018testing,liang2024structural}, generating a null‐model of regional $t$‐values. Positive $z$‐scores denote \emph{accelerated tau deposition} relative to the null model. Asterisks indicate $p_{\mathrm{spin}}<0.05$ (e.g.\ Temporal lobe: $p=0.024$, Subcortical: $p<10^{-5}$).  
SUVRs mapped in ParaView (v5.10.1) \footnotemark[2].  
}
\label{fig1}
\end{figure}

\textbf{Data description.} To investigate the relationship between tau propagation and both the functional and structural connectomes during the progression of AD, we leveraged multi-modal neuroimaging data, including tau PET, resting-state fMRI, and diffusion MRI, from two longitudinal cohort comprising 839 brain scans from 539 participants (aged 55–94 years $75.06\pm 7.70$, female: male= 278:261) in the Alzheimer’s Disease Neuroimaging Initiative (ADNI cohort). Among the participants, 329 underwent a single scan, 136 underwent two scans separated by an average interval of 1.25 years, and 60 underwent three scans separated by an average interval of 1.11 years, 12 underwent four scans separated by an average interval of 1.11 years, 2 underwent five scans separated by an average interval of 0.92 years. Baseline diagnostic categories were: Alzheimer’s disease (AD, $N$ = 54, 85 scans in total, 55y-94y), cognitively normal (CN, $N$ = 157, 237 scans in total, 56y-94y), early mild cognitive impairment (EMCI, $N$ = 117, 186 scans in total, 58y-93y), late mild cognitive impairment (LMCI, $N$ = 60, 105 scans in total, 56y-92y), and subjective memory complaints (SMC, $N$ = 145, 210 scans in total, 57y-90y); diagnosis was unavailable for 14 participants (16 scans, 58y-85y). Fig. \ref{fig1}a illustrates the longitudinal follow-up of participants with more than one scan, alongside the changes in average regional tau standardized uptake value ratio (SUVR) stratified by clinical labels. Participants were stratified into two groups based on clinical diagnosis (disease severity): `CN'-group, comprising CN, SMC, and EMCI participants; and `AD'-group, comprising LMCI and AD participants. This binary grouping provided the framework for subsequent group‐level analyses.

\textbf{Regional evolution of tau PET uptake from CN to AD.} For each participant, the brain was parcellated into 148 cortical and 12 subcortical regions using the Destrieux atlas \cite{destrieux2010automatic}. We then derived regional tau SUVR from positron emission tomography (PET) (see Appendix \ref{data-pre} for full image pre-processing details).

To identify the spatial pattern of tau propagation, we first examined the regional differences of SUVR between the CN and AD groups via a mixed‐effects linear model \cite{laird1982random}, including sex as a covariate. Significant increases in tau deposition (Bonferroni-corrected $p<0.05$, $t$-value$>4.0$) were primarily detected in the dorsolateral prefrontal cortex, lateral temporal cortex, and medial temporal pole/entorhinal cortex (Fig.~\ref{fig1}b).

\footnotetext[1]{\url{https://brainspace.readthedocs.io/en/latest/pages/matlab_doc/main_functionality/spin_permutations.html}}

\footnotetext[2]{\url{https://www.paraview.org/}}

Next, we grouped the 148 cortical and 12 subcortical nodes into six anatomical lobes (frontal, insula, temporal, occipital, parietal, limbic) \cite{destrieux2010automatic} plus subcortex, and performed 1,000 spin‐permutation tests to correct for spatial autocorrelation \cite{alexander2018testing,liang2024structural} to assess whether tau propagation concentrates in specific large‐scale systems. We then computed each lobe’s average $t$–value and expressed it as a $z$–score relative to the null distribution. Although raw SUVR increases were observed across all lobes, only the temporal lobe ($p_{\mathrm{spin}}=0.024$) and subcortical regions ($p_{\mathrm{spin}}<10^{-5}$) showed significantly accelerated tau aggregation. The other lobes did not exceed—and in some cases fell below—their null expectations (Fig. \ref{fig1}c). We then mapped each of the 160 regions onto the 13 canonical functional modules of Power et al. \cite{power2011functional} and re‐ran the spin‐permutation test (Fig.~\ref{fig1}d). While most networks showed $z$-scores at or below zero once spatial autocorrelation was accounted for, the subcortical module again emerged as a pronounced area that is significantly associated with tau increase ($p_{\mathrm{spin}}<10^{-5}$). No other functional system exceeded its null‐model expectation, even those with modest SUVR increases. This confirms that tau accumulation from CN to AD is not evenly distributed across large‐scale functional networks but is instead highly concentrated in subcortical circuits only. 
Together, our analyses demonstrate that temporal lobe and subcortical regions exhibit accelerated tau deposition, indicating their role as key epicenters that anchor the connectome-mediated propagation of disease.

\subsection{Supporting evidence I: Tau propagation is shaped by structural and functional networks}
\label{sec_supportI}
In this section, we examine whether regional patterns of tau accumulation underlie the brain’s intrinsic connectivity architecture. 
In this work, we parcelate brain into 148 cortical regions and 12 sub-cortical regions, yielding a $160 \times 160$ structural network (derived from diffusion‐weighted MRI tractography) and a corresponding $160 \times 160$ functional network (derived from region-to-region correlation of blood-oxygen-level dependent (BOLD) signals from resting‐state fMRI). The group average \footnote{For simplification, element-wise average is used to demonstrate the average of brain networks across individuals.} of SC and FC are shown in Fig. \ref{fig2}a. Full details of the SC and FC preprocessing pipelines are provided in Appendix \ref{data-pre}. 

First, we employed a mixed-effects model to predict diagnostic labels based on longitudinal tau SUVR measurements at each brain region, where age, sex, and \textit{APOE4} status are confounders. This analysis yielded a regional tau extent, denoted as $T_i$, represented by the $t$-value from the model. Since $T_i$ characterize the effect size of tau aggregates on clinical outcome, it is less influenced by inter-subject variability in tau SUVR. In this regard, we focus on tau extent $T_i$, instead of original tau SUVR, in the following network autocorrelation test to examine whether AD-specific change of tau aggregates is shaped by SC or FC \footnote{The network autocorrelation test result on the change ratio of tau SUVR is shown in Fig. \ref{fig3}e.}.
To do so, we evaluate whether AD-related tau accumulation at a given brain region can be predicted from the tau changes observed in its anatomically or functionally connected neighboring regions, which is an essential premise underlying the connectome-based tau propagation framework.
Specifically, the statistical analysis involves two main steps: (1) For the $i^{th}$ brain region, calculate the mean tau extent of its directly connected neighbors through SC $\hat{T}_i^S=\sum_{j\in \mathcal{N}_S(i)}T_j$ and FC $\hat{T}_i^F=\sum_{j\in \mathcal{N}_F(i)}T_j$ respectively, as demonstrated in Fig. \ref{fig2}b, where $\mathcal{N}_S(i)$ and $\mathcal{N}_F(i)$ are SC- and FC-based graph neighborhood of the $i^{th}$ brain region, respectively. 
(2) examine the correlations of $T_i \sim \hat{T}_i^{S}$ and $T_i \sim \hat{T}_i^{F}$. 
A strong correlation indicates that the spatial pattern of tau burden is closely shaped by the brain’s intrinsic connectivity architecture, suggesting that tau propagation likely follows a network-based spreading mechanism. Otherwise, tau burden is randomly distributed. 

Across all nodes, both $\hat{T}_i^S$ and $\hat{T}_i^F$ were strongly predictive of the underlying region’s own tau extent $T_i$ (Fig. \ref{fig2}c, SC: adjusted $r=0.783$, $p<0.001$, FC: $r=0.736$, $p<0.001$, two‐sided), indicating that the increase of tau aggregates from CN to AD is associated with the topology of SC and FC. To ensure these relationships were not driven by spatial autocorrelation, we aggregate nodes into seven anatomical lobes and ran 1,000 spin‐permutation tests on the lobe‐wise SC/FC correlations, controlling FDR at $p_{\text{spin}=0.01},$ one‐tailed). Only multi-sensory areas—particularly frontal, temporal, and parietal cortices—and subcortical circuits exhibited correlations that significantly exceeded their null expectations (Fig.~\ref{fig2}d). This pattern confirms that tau spread preferentially follows both white-matter tracts and functional pathways within higher-order, multi-sensory and subcortical networks, in line with established Braak staging and PET findings \cite{braak1991neuropathological,jack2018nia}.

We then investigate whether the longitudinal change of tau pathology follows FC or SC networks. To do so, we conduct node-wise multiple linear regression analyses, separately modeling FC and SC measures (nodal connectivity degree) to predict regional tau accumulation rates ($\Delta$Tau) and cognitive performance (MMSE), controlling for age, sex, and APOE genotype (Fig. \ref{fig2}e–f). Our analyses revealed distinct connectivity patterns: FC predominantly explained regional variance in tau propagation, implying functional networks as primary pathways facilitating the spread of tau pathology. In contrast, SC was more strongly associated with cognitive performance, suggesting that intact structural networks are critical for supporting cognitive resilience. These findings underscore dissociable roles of brain connectivity modalities in AD progression, i.e., functional networks primarily mediate tau pathology spread through neuronal activity-dependent mechanisms, whereas structural networks serve as essential anatomical substrates underpinning cognitive reserve. This aligns closely with current neuroimaging evidence \cite{franzmeier2020functional,vogel2020spread,van2019cross} and contributes to our understanding of the complementary yet distinct roles of functional and structural networks in AD pathology and cognition.

\begin{figure}[h!]
\centering
\includegraphics[width=1\textwidth, trim=0 80 0 0, clip]{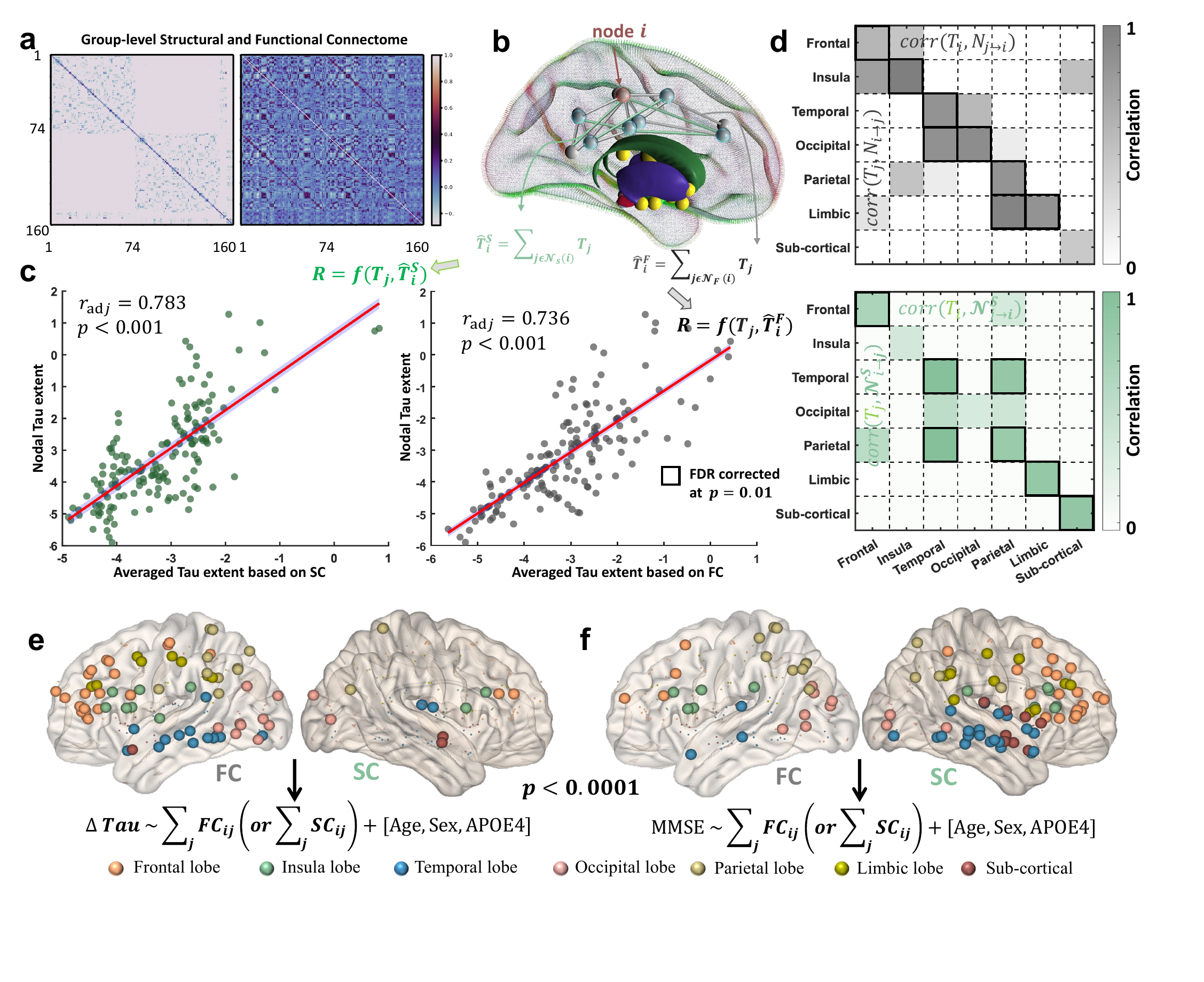}
\caption{\textbf{Tau propagation is anchored to structural and functional network backbones.} \textbf{(a)}. Group‐average structural (left) and functional (right) connectivity matrices ($160\times 160$) thresholded to their backbone edges. \textbf{(b)}. Schematic illustrating how we relate regional tau propagation to network structure: (1) for each node $i$ (red), we compute the tau extent $T_i$ ($t$-value derived from mix-effect model of predicting diagnostic label using longitudinal regional tau SUVR); (2) compute averaged tau extent of its directly connected neighbors $\hat{T}_i^S$ based on SC (green edges) and $\hat{T}_i^F$ based on FC (gray edges), then correlate $\hat{T}_i^S\sim T_i$ and $\hat{T}_i^F\sim T_i$, respectively. 
\textbf{(c)}. $T_i\sim \hat{T}_i^S$ (left) and $T_i\sim \hat{T}_i^F$ (right) across all nodes reveals strong positive associations 
(Pearson correlation, SC: $r{_{\text{adj}}=0.783, p<0.01}$, FC: $r{_{\text{adj}}=0.736, p<0.01}$, two sided). Lines show least‐squares fits; shaded areas denote 95\% confidence intervals. 
\textbf{(d)}. Lobe‐level spin‐permutation testing (FDR corrected at $p{_{\text{spin}}=0.01},$ one‐tailed) confirms that only multi-sensory area (frontal, temporal, and parietal)  and subcortical systems exhibit connectivity–tau correlations exceeding spatial‐null expectations (colored boxes). Multiple linear regression analyses of FC and SC associations with \textbf{(e)} tau accumulation rate ($\Delta$Tau) and \textbf{(f)} MMSE, respectively. Both models include age, sex, and APOE genotype as covariates to control for potential confounding effects. 
Our analyses revealed that FC primarily drives tau propagation, while SC is more closely linked to cognitive performance, indicating distinct roles in disease spread and resilience.
 }
 \label{fig2}
\end{figure}

\subsection{Supporting evidence II: Tau aggregation progresses through multiple distinct stages}
\label{sec_supportII}

\begin{figure}[h!]
\centering
\includegraphics[width=1\textwidth]{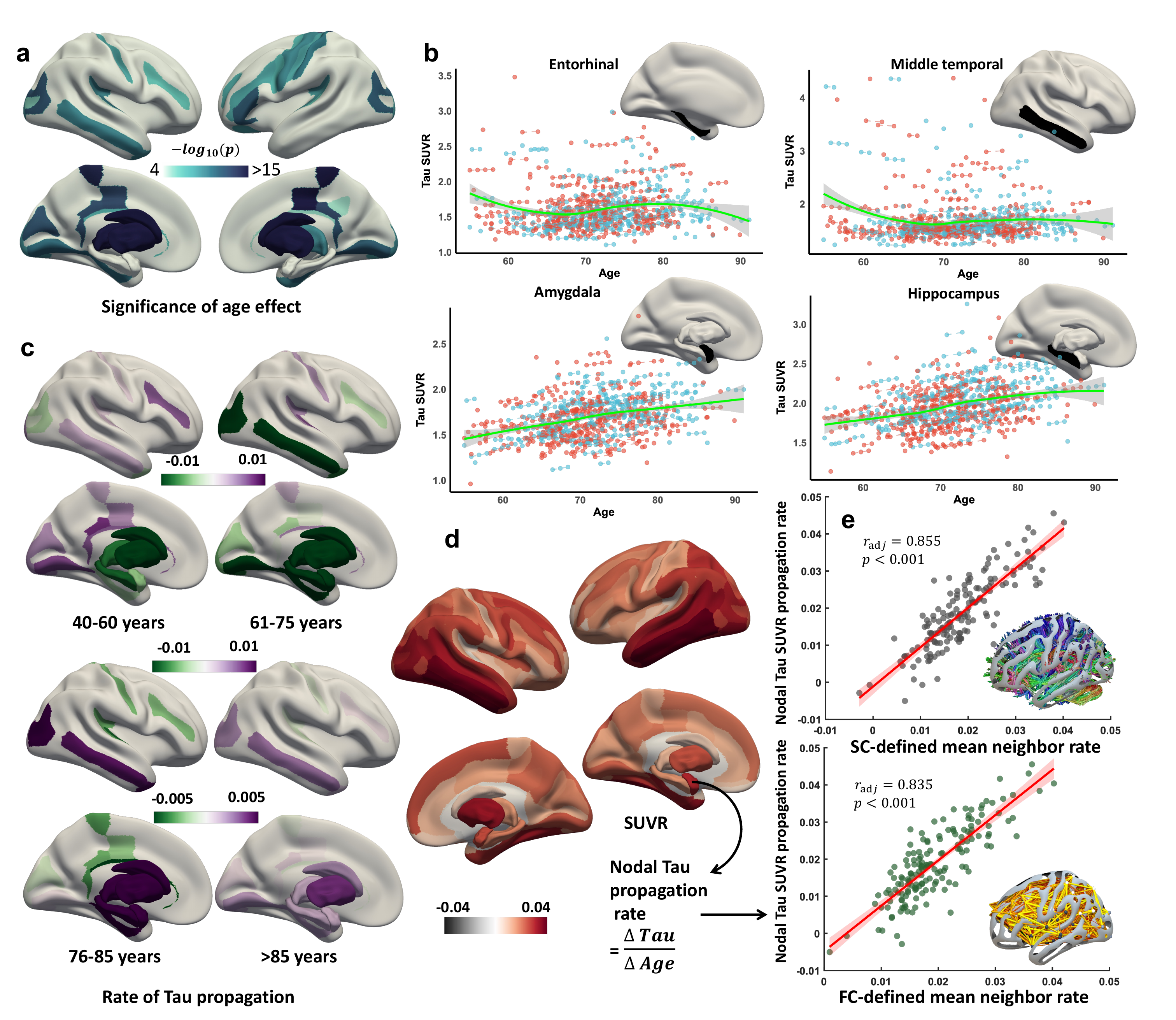}
\caption{\textbf{Age‐related and individual‐level analyses of tau propagation along SC and FC.}  
\textbf{Panels a–c: GAM with age as a continuous predictor.}  
\textbf{(a)} Cortical surface map of the effect size of age on tau accumulation, Bonferroni‐corrected at $p<5\times10^{-5}$, two‐sided.  
\textbf{(b)} Example GAM fits for entorhinal cortex, middle temporal gyrus, amygdala, and hippocampus: green lines denote the estimated age‐trajectory of tau SUVR; shaded gray bands show 95\% confidence intervals.  
\textbf{(c)} Spatial maps of the first derivative of the GAM fit (tau propagation rate) plotted for four age windows: $<$60, 61–75, 76–85, and $>$85 years.  
\textbf{Panels d–e: individual‐level longitudinal analysis.}  
\textbf{(d)} Illustration of how nodal tau propagation rates are computed from each subject’s serial scans. ($\Delta$Tau/$\Delta$Age), and the resulting group‐mean surface map (averaged over all 210 participants) showing the spatial distribution of propagation velocities—highest in medial temporal and subcortical regions. \textbf{(e)} Scatter plot of nodal rate versus neighbor‐mean rate in SC (gray) and FC (green), with linear fits and 95\% confidence intervals (red shading). Reported correlations: SC $r_{\rm adj}=0.855$, FC $r_{\rm adj}=0.835$, both $p<0.001$, two‐sided.    
}
\label{fig3}
\end{figure}

To determine whether connectivity‐guided tau spread persists across the adult lifespan, we conducted two complementary analyses (Fig. \ref{fig3}): a continuous‐age generalized additive model (GAM) \cite{wood2004stable} and an individual‐level longitudinal analysis.

\textbf{GAM‐estimated age effects}:  We fit a GAM in which chronological age serves as a smooth predictor of tau extent $T_i$ for each brain region. After Bonferroni correction at $p<5\times10^{-5}$ (two‐sided), we mapped the age effect on tau extent onto the cortical surface (Fig.~\ref{fig3}a).  The resulting $-\log_{10}(p)$ map highlights that \textit{medial temporal} (entorhinal, parahippocampal) and \textit{sub-cortical} (hippocampus, amygdala, caudate and putamen) regions exhibit the strongest positive associations between age and tau$\sim$(disease) phase relationship, indicating that aging might escalate the effect of tau aggregation on disease progression in these identified areas.

\textbf{Regional GAM trajectories:} To illustrate the non‐linear age trajectories captured by the GAM, we plot the SUVR data points and the fitted age curves (green) with 95 \% confidence intervals (gray shading) for four key regions in tau propagation: \textit{entorhinal cortex}, \textit{middle temporal gyrus}, \textit{amygdala}, and \textit{hippocampus} (Fig.~\ref{fig3}b). Each curve represents region‐specific dynamics:  

- The \textit{entorhinal cortex} shows an early plateau or slight mid‐life decline (ages 60–75) followed by renewed late‐age acceleration—mirroring Braak I/II progression \cite{braak1991neuropathological}. 

- The \textit{middle temporal} attains a high plateau in mid‐life and then maintains only gradual increases through older age, consistent with Braak III/IV staging.  

- The \textit{amygdala} exhibits a near‐monotonic, approximately linear rise in SUVR across the entire age span, reflecting steady limbic tau buildup.  

- The \textit{hippocampus} likewise demonstrates continuous upward drift, with particularly steep increases in mid‐ to late life and no late‐age downturn, as expected for this early‐affected region.

\textbf{Age‐window propagation rates:} 
By fitting the relationship between regional tau accumulation and age using GAM for each brain region, we have identified four distinct stages based on \textit{tau propagation rates} ($\frac{\Delta Tau}{\Delta age}$) as: $<$60, [61–75], [76–85], and $>$85 years (Fig.~\ref{fig3}c).  
The spatiotemporal pattern of tau pathology in each stage is summarized below:  

- \textit{$<$60 years}: Propagation concentrates in neocortical association zones (lateral temporal, inferior parietal), while medial temporal and subcortical rates remain modest. 

- \textit{61–75 years}: Spread intensifies within heteromodal hubs \footnote{Network nodes with exceptionally high connectivity that integrate information across multiple functional systems. These hubs not only exhibit dense structural links but also strong functional coupling with diverse brain regions, facilitating coordination and information flow. Typical examples are the posterior cingulate cortex and dorsolateral prefrontal cortex.} (middle temporal, temporal pole) and begins encroaching on limbic structures (amygdala, parahippocampal gyrus), in line with Braak III/IV \cite{braak1991neuropathological}.  

- \textit{76–85 years}: The focus shifts to medial temporal and subcortical circuits (entorhinal cortex, hippocampus, striatum (caudate nucleus and putamen)), reflecting advanced Braak V/VI pathology \cite{braak1991neuropathological}. 

- \textit{$>$85 years}: Overall rates diminish, yet subcortical regions retain the highest residual propagation velocity, indicating they remain the last loci of active tau accumulation in the oldest individuals.
 
\textbf{Brain mapping of tau propagation rate.} To assess tau propagation profile across individuals, we calculate each subject’s nodal tau propagation rate ($\Delta$Tau/$\Delta$Age) between consecutive PET visits.  We then average these rates at each node across all 210 participants and project the resulting group‐mean rate map onto the cortical and subcortical surfaces (Fig.~\ref{fig3}d). This group‐average map highlights that, collectively, medial temporal and subcortical regions exhibit the highest tau propagation velocities, closely matching the individual‐level and continuous‐age analyses.

\textbf{Tau propagation rate is shaped by both SC and FC.} For every individual, we correlated their nodal propagation rates with the averaged rate of directly connected neighbors in SC and FC (scatter plots, Fig.~\ref{fig3}e).  At a significant level $p<0.001$, SC‐neighbor and FC‐neighbor rates explain 85.5 \% and 83.5 \% of the variance in nodal rates, respectively. 
Across all participants, the adjusted R-squared $r_{\rm adj}$ ranges from $0.679\sim 0.855$ for SC-based and $0.622\sim 0.835$ for FC-based analysis ($p<0.001$ after spinning test). 


\subsection{Explore novel mechanism in connectome-based Tau propagation hypothesis}
\label{sec_main}

To investigate the respective and combined influences of structural and functional connectivity on tau propagation, we developed a multi‐layer transport framework (detailed in Sec. \ref{sec:methods}) that jointly models SC and FC pathways to capture the spatiotemporal spread of tau SUVR from longitudinal PET imaging \cite{huang2025multilayer}. Specifically, we integrate SC and FC graphs into a unified diffusion‐reaction operator where the SC-FC interaction are learned via a physics‐informed neural network (PINN), enabling us to evaluate propagation rates directly from baseline scans and forecast future tau accumulation. Our multi-layer model has achieved promising accuracy in predicting future tau aggregation with a prediction error of less than 6.2\%, indicating the potential of our computational approach for uncovering novel mechanisms in tau propagation.
Full details of the model formulation, training procedure, and evaluation protocols are provided in the Methods section \ref{sec:methods}.
In addition, we disentangle the SC-specific tau propagation as $u_s$ and the FC-specific tau propagation $u_f$, which allows us to explore how anatomical wiring and functional co‐activation jointly shape pathology spread by examining the following hypotheses.

\begin{figure}[h!]
\centering
\includegraphics[width=0.9\textwidth, trim={0cm 0cm 0cm 0cm}, clip]{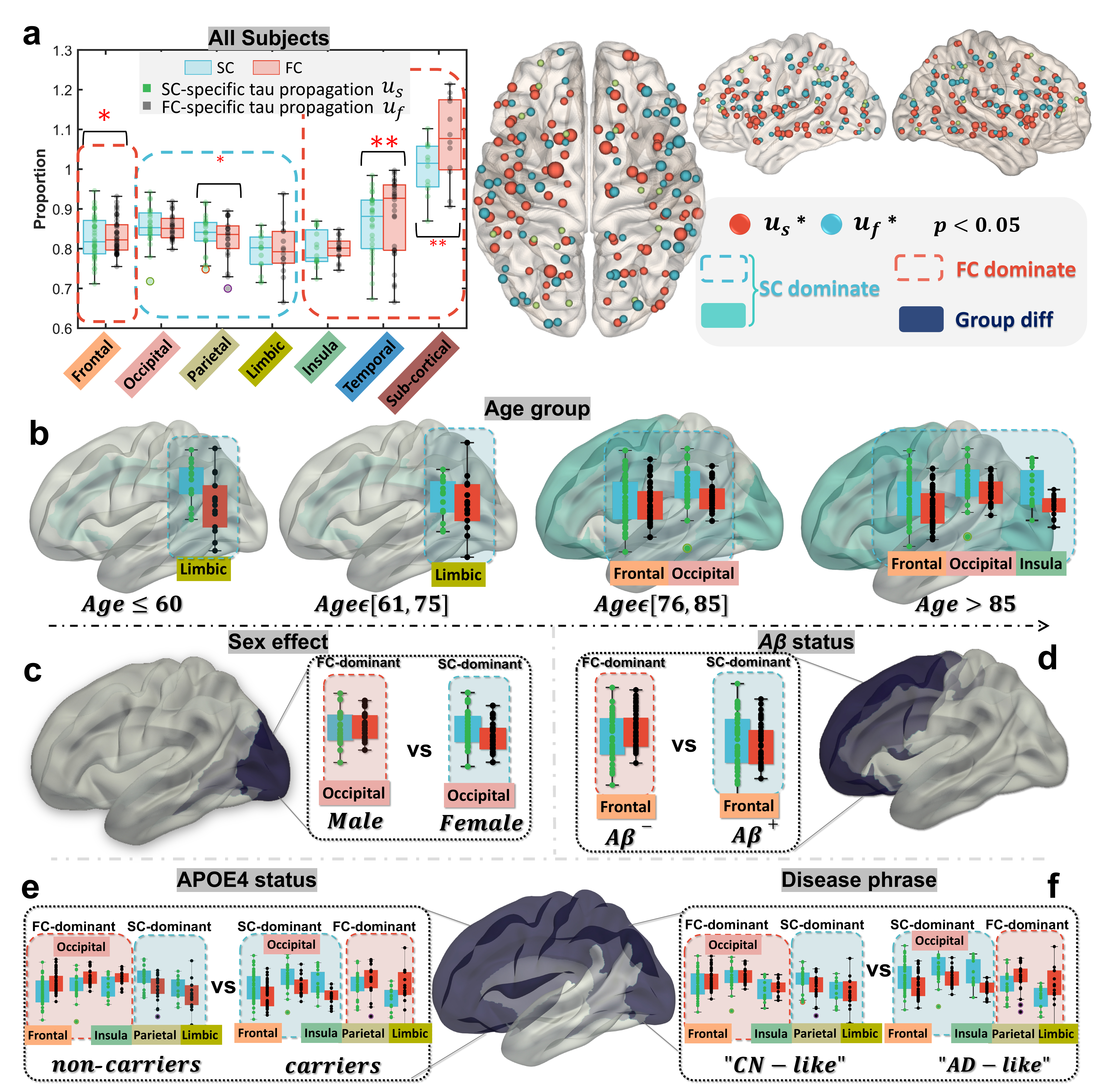}
\caption{\textbf{Mechanistic role of SC and FC on tau propagation.} \textbf{(a) Main findings.} \textit{Left}: TProportions of tau propagation attributed to structural connectivity ($u_s$, blue) and functional connectivity ($u_f$, red) across different brain lobes.  \textit{Right}: Comparative mapping of $u_s$ versus $u_f$, where blue marks regions with stronger SC‐driven propagation and red highlights those with stronger FC‐driven propagation. Node sizes scale with the absolute difference in magnitude, $|u_s - u_f|$. \textbf{(b) Dynamic contribution of SC and FC to tau propagation.} SC and FC contributions to tau propagation across four age stages: $<$60, 61–75, 76–85, and $>$85 years. Age-dependent shift in tau propagation, with younger individuals showing FC-dominant spread and older individuals showing increasing reliance on SC pathways.
\textbf{(c) Stratification analysis by biological sex.} Sex-dependent effect only manifests a minor modulatory effect on tau propagation, with a modest occipital difference that does not alter the overall SC-to-FC propagation architecture.
\textbf{(d) Impact of amyloid-$\beta$ deposition. } SC vs. FC comparisons between $A\beta+$ and $A\beta-$ individuals suggest that A$\beta$ burden “boosts” tau spread along functional circuits (especially in frontal cortex). 
\textbf{(e) The effect of \textit{APOE4} status on region-specific network conduit in tau propagation.} \textit{APOE4} carriers show a transition from FC- to SC-dominant tau spread in the frontal, occipital, and insula cortex.
\textbf{(f) Stratified results on disease phrase.} Group-wise propagation patterns across clinical diagnoses (e.g., CN, AD). Except for the temporal and subcortical regions, all other lobes exhibited a reversal in SC–FC dominance between AD and CN groups.
. 
}\label{fig_main}
\end{figure}

\textbf{Hypothesis I: The mechanistic role of SC and FC in tau propagation might vary across different brain regions} 
In the left panel of Fig. \ref{fig_main}a, we show the bar plot of the disentangled volumes of tau aggregation associated with SC ($u_{s}$ in blue) and FC ($u_f$ in red), across brain lobes, using all ADNI subjects that have longitudinal tau PET scans. For each brain region, we examine whether SC leads the predominant role in local tau propagation (i.e., $u_s > u_f$ at a significant level of $p<0.05$) or vice versa, using a one-sided paired t-test. In the right panel of Fig. \ref{fig_main}a, brain regions where tau propagation is primarily driven by SC and FC are shown in blue and red, respectively, with node size reflecting the magnitude of the difference between $u_s$ and $u_f$.

\textit{Discussion.} The results shown in Fig. \ref{fig_main}a indicate the following findings. 
(1) Tau propagation at association cortices \footnote{High‑order cortical regions outside primary sensory and motor areas that integrate multimodal information (e.g., visual, auditory, somatosensory) and support complex cognitive functions such as attention, memory, language, and decision‑making. Examples include the prefrontal cortex, posterior parietal cortex, and superior temporal sulcus.} and subcortical hubs is primarily driven by FC. Specifically, the frontal, insula, temporal, and subcortical regions all exhibit median FC-fractions of ~60–70\%, well above chance. This indicates that, over a short-term evolution (e.g., 1-2 year), tau accumulation in these areas follows patterns of synchronous activity (“who-fires-together”) more than anatomical fiber tracts. 
(2) Posterior and paralimbic cortices remain tract-anchored. By contrast, the occipital, parietal, and limbic lobes each show median SC-fractions just above 50\%, revealing that one-year tau increases here still rely more on white-matter pathways than on functional co-activation. 

\textbf{Hypothesis II: Tau propagation might be shaped by time-varying contributions from both SC and FC.}
To investigate age-related shifts in the spatial patterns of SC-dominant and FC-dominant tau propagation, we stratified participants into four age groups as defined in Fig. \ref{fig3}c and applied the same statistical framework used to test \textbf{Hypothesis I} within each subgroup. As illustrated in Fig. \ref{fig_main}b, younger individuals (age $<60$ years) exhibited a spatial pattern in which tau spread was primarily driven by FC, with dominant effects observed in the frontal, temporal, and subcortical regions. However, this pattern shifted progressively with age. In individuals aged 61–75 years, structural connectivity began to play an increasingly prominent role, particularly in the frontal and occipital cortices. This trend continued into older age groups: among participants aged $76–85$ years, SC-driven propagation extended into the insular and limbic cortices. In the oldest group (age $>85$ years), SC dominance became widespread, with structural pathways emerging as the principal conduits for tau diffusion across nearly several cortical lobes. 

\textit{Discussion.} These findings suggest a dynamic shift in the mechanisms underlying tau spread in the aging brain, whereby functional pathways dominate in early stages and structural pathways progressively assume control as the brain ages, which are consistent with prior studies showing that tau initially spreads through functional connectivity but progressively depends more on anatomical pathways as aging or disease progresses \cite{franzmeier2019functional, alexander2018testing}.
The evolving mechanism of tau spread has important implications for both early intervention and disease prevention in AD. In preclinical or prodromal stages when tau is likely disseminating via functional networks, interventions that target network hyperactivity \cite{penalba2023increased}, synaptic dysfunction \cite{shankar2009synaptic}, or excitatory-inhibitory imbalance \cite{wang2025eiimbalance} may be especially effective in halting or slowing the initiation of tau propagation. Functional imaging biomarkers, such as resting-state fMRI or MEG-based connectivity, could therefore play a crucial role in identifying individuals at risk and monitoring early treatment response.
As the disease advances and structural pathways begin to dominate, therapeutic strategies may need to pivot toward preserving axonal integrity, promoting white matter health, and enhancing neurovascular support. For instance, treatments aimed at reducing myelin degradation \cite{depp2023myelin}, maintaining cytoskeletal stability \cite{brunden2020epoD}, or improving glymphatic clearance \cite{murdock2024multisensory} might be more effective at this stage. This mechanistic shift also implies the value of longitudinal, multimodal imaging to capture the dynamic interplay between functional and structural connectivity, enabling stage-specific therapeutic targeting.

\textbf{Hypothesis III: The mechanistic role of SC and FC in tau propagation might be altered by non-modified AD risk factors.}
\textit{APOE4} is one of the strongest genetic AD risk factors. Meanwhile, accumulating evidence indicates that AD manifests with sex-specific differences, with a higher prevalence reported in females. To examine how these biological factors influence the mechanisms of tau propagation, we stratified our analysis by (1) \textit{APOE4} carrier status \footnote{\textit{APOE4} carrier indicates carrying either one copy or two copies of $\epsilon$4.} (Fig. \ref{fig_main}e) and (2) biological sex (Fig. \ref{fig_main}c), and assessed the regional dominance of SC versus FC in driving tau spread.
Sex-stratified analyses revealed largely consistent SC–FC dominance patterns across most brain regions between males and females. However, a notable divergence emerged in the occipital cortex, where males exhibited a shift toward FC-dominant tau propagation. This localized difference may reflect sex-specific vulnerabilities in posterior functional networks or divergent neurodegenerative trajectories across sexes, which aligns with prior reports of sex-based distinctions in brain connectivity and disease progression trajectories \cite{ferretti2018sex, satterthwaite2014sex}.
By contrast, \textit{APOE4} carrier status exerted a more substantial effect on the regional dynamics of tau propagation. In \textit{APOE4} carriers, we observed a pronounced shift from FC- to SC-dominant tau spread in the frontal, occipital, and insular cortices—potentially reflecting a compensatory reliance on preserved structural pathways amid functional network disruption. Conversely, the parietal and limbic regions exhibited a transition from SC- to FC-dominance, indicating increased vulnerability of these heteromodal and memory-related systems to network-based propagation. This reconfiguration was widespread: all lobes except the temporal and subcortical regions demonstrated a reversal in SC–FC dominance, suggesting a global restructuring of propagation mechanisms in individuals with elevated genetic risk. These findings align with prior evidence linking \textit{APOE4} to early breakdown of default mode network hubs and accelerated cortical disintegration, which may shape the evolving pathways through which tau pathology spreads across disease stages \cite{jagust2018imaging}. Full statistical details and additional results are provided in Fig. \ref{adniscfc3} of Appendix \ref{supp_results}. 

\textit{Discussion.} 
The effect of \textit{APOE4} status on tau propagation appears to be heterogeneous. The transition toward SC-dominance in frontal and insular cortices among \textit{APOE4} carriers suggests that individuals with elevated genetic risk may engage alternate anatomical routes for pathological spread, which may represent a form of structural compensation potentially due to early impairment of functional integrity. 
Meanwhile, the observed shift toward FC-dominance in parietal and limbic regions may reflect increased vulnerability of association networks central to memory and cognitive control.

Given that APOE, especially the APOE $\epsilon$4 allele, is crucial for lipid transport and metabolism and its function is modulated by sex hormones, carrying \textit{APOE4} confers a greater risk for AD in females than in males \cite{altmann2014sex}. In light of this, we further investigate the sex-by-\textit{APOE4} effect on the tau propagation volume separately driven by SC, FC, and their difference, alongside the main effects of sex and \textit{APOE4} individually. At a significant level of $p<0.05$, none of the brain regions manifests a significant joint effect within the sensitivity limits of our cohort and multi-layer modeling framework. 

Furthermore, we stratified subjects into ``CN-like" (CN, SMC, EMCI) and ``AD-like" (LMCI, AD) phases based on disease severity reflected in the diagnostic labels and investigated the contribution of SC and FC in tau propagation. As shown in Fig. \ref{fig_main}f, the outcome of disease progression exhibits very similar spatial patterns of SC$\leftrightarrow$FC transition as the effect of \textit{APOE4} shown in Fig. \ref{fig_main}e. That is, a shift from FC- to SC-dominant tau propagation in the frontal, occipital, and insular cortices, while the parietal and limbic regions exhibited a transition from SC- to FC-dominance.

\textbf{Hypothesis IV: Excessive amyloid accumulation may shift the balance between SC and FC in driving tau propagation.}
We stratified subjects based on A$\beta$ status using a threshold of 192 pg/mL (A$\beta^+$: $<$192; A$\beta^-$: $\ge$192) on CSF biomarker. As shown in Fig. \ref{fig_main}d, the increase of A$\beta$ burden at the frontal cortex is associated with the shift of connectome driving factors from FC to SC. While A$\beta$ deposition does not substantially alter the overall dominance pattern between SC and FC, it appears to selectively modulate this balance in specific regions, most notably the frontal cortex \cite{nabizadeh2024abeta}. 

\textit{Discussion.} One possible interpretation for the observed transition from FC‐dominant to SC‐dominant tau propagation in frontal cortex with elevated amyloid plaque accumulation is as follows:  
Multiple threads of evidence show that A$\beta$ preferentially targets highly interconnected functional hubs, especially within the default mode and frontoparietal control networks, leading to early synaptic dysfunction and disintegration of functional coherence \cite{buckner2005default, palop2007network, jagust2018imaging}. In this context, FC degradation may precede and facilitate tau spread by weakening activity-dependent clearance mechanisms and disrupting the propagation of neuronal signals that normally constrain pathological spread. As functional networks lose integrity under the toxic influence of A$\beta$, the brain may increasingly rely on residual anatomical pathways to support the transneuronal transmission of tau aggregates. This compensatory shift from FC to SC dominance mirrors recent evidence showing that functional vulnerability is an early hallmark of A$\beta$-related pathology, while structural degeneration becomes more pronounced as the disease progresses \cite{seeley2009neurodegenerative, franzmeier2020functional}. 
Importantly, the selective modulation of tau propagation by A$\beta$ burden supports the view that amyloid and tau act synergistically yet distinctly on brain networks, and that A$\beta$ may not only trigger tau pathology but also reshape its preferred routes of spread. Understanding how regional A$\beta$ deposition alters the balance between functional and structural propagation pathways may inform more targeted therapeutic interventions aimed at preserving network integrity and slowing disease progression.

\textbf{Hypothesis V: The spatial pattern of SC/FC dominance in tau propagation might be linked to genetic risk factors.}
To identify molecular drivers of spatial SC-dominant and FC-dominant patterns in tau propagation, we performed a weighted gene co-expression network analysis (WGCNA) \cite{langfelder2008wgcna} and functional enrichment analysis from 15,633 genes. Specifically, we first retrieved 764 AD-related genes from the MalaCards Human Disease Database \cite{Rappaport2017MalaCards}. On top of these, 579 genes were mapped to the Abagen-derived \cite{markelloabagen} expression matrix for $N=160$ Destrieux parcels (see the Appendix \ref{gene_exp}, Table \ref{label_name}) and used for WGCNA. This analysis identified a key gene module (``salmon" module) containing 18 AD-related genes---including RCAN1, NOTCH1, PICALM, ADAM10, HSPD1, NFE2L2, HSPA5, MCL1, MAP2K7, CHUK, IKBKB, CTNNB1, RHOA, SIRT1, SP1, TH, TMEM106B, and CSNK1A1 (resulting in $G=18\times N$ gene expression data matrix)---that was preferentially expressed in 12 subcortical regions. Further analysis of this module, including gene expression specificity, co-expression correlations, and protein-protein interaction networks, is presented in Appendix \ref{gene_exp}.

Next, we determined which risk genes contribute to SC/FC-dominant brain mapping in tau propagation. To do so, we applied non-negative least absolute shrinkage and selection operator (LASSO) regression with $B = 100$ bootstrap resamples. First, we conducted group comparisons between AD and CN subjects for both $u_s$ and $u_f$ components using mixed-effects modeling to identify significant cortical regions. The resulting two-sided $p$-values were transformed into $-\log_{10}p$ scores and used as the target variable in LASSO regression, i.e., $\hat{\beta} = \arg\min \left\{ \frac{1}{2n} \| -\log_{10}({p}) - G\beta \|_2^2 + \lambda \| \beta \|_1 \right\}$. These scores were regressed against the $N \times G$ gene expression matrix $G$, and the frequency $s_j$ of each gene $j$ being selected with a nonzero coefficient across resamples was recorded:$s_j = \frac{1}{B} \sum_{b=1}^{B} \mathbb{I}\left( \beta_j^{(b)} \neq 0 \right)$. Here, $\lambda$ is a regularization parameter that controls sparsity in the regression model: higher values of $\lambda$ enforce greater sparsity by penalizing large coefficients, leading to fewer genes being selected. We explored a logarithmic grid of 15 values ranging from 0.0001 to 1 to determine robust gene signatures. The vector $\beta$ in the LASSO regression represents the regression coefficients assigned to each gene. Each element $\beta_j$ quantifies how strongly the expression of gene $j$ contributes to predicting the regional sensitivity to SC–FC propagation asymmetry (measured by $-\log_{10}p$). If $\beta_j = 0$, gene $j$ is not predictive (excluded by LASSO). If $\beta_j \ne 0$, gene j contributes to the model and is considered a candidate biomarker (i.e., $\mathbb{I}$ is an indicator function). Genes with consistently high selection frequency ($s_j$) across bootstraps were considered stable predictors.


\begin{figure}[ht]
\centering
\includegraphics[width=\textwidth, trim={0cm 0cm 0cm 3cm}, clip]{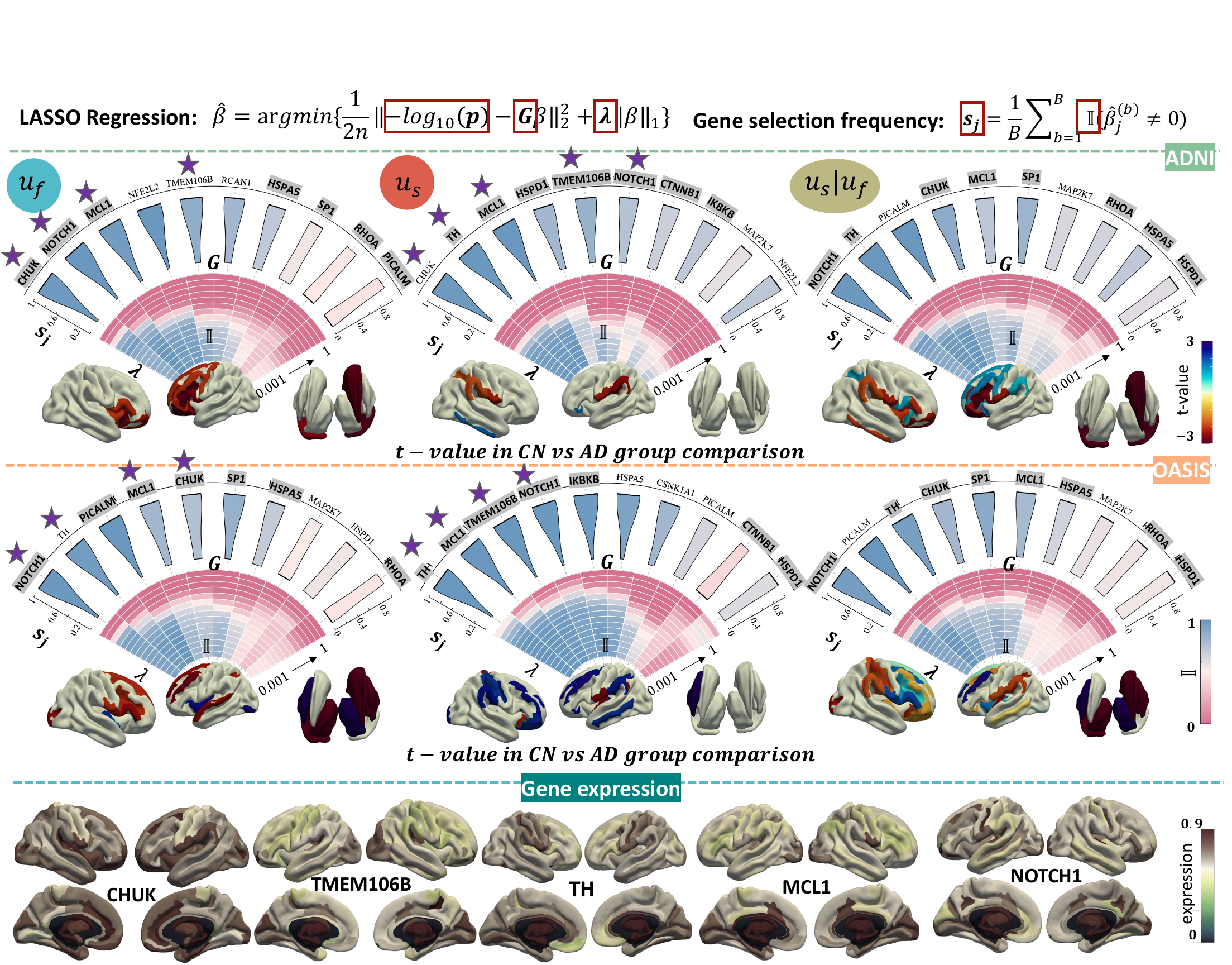}

\caption{\textbf{Gene correlates of SC- and FC-dominant tau propagation on ADNI and OASIS datasets.}  
\textbf{Top:} LASSO selection results using tau propagation associated with FC ($u_f$) (left), associated with SC ($u_s$) (middle), and the selection pattern of SC-FC dominance that both $u_s$ and $u_f$ exhibit significant difference between CN and AD (right). Pie chart--selection paths across regularization ($\lambda$) for the top 10 genes (first six genes are more stable). Violin diagram: Gene selection frequency over 100 non-negative LASSO bootstrap resamples.  
Brain mapping under the violin diagram: $t$-value in each CN vs AD group comparison ($p<0.05$).
Shaded genes are the common genes on both ADNI and OASIS datasets. Star denotes the common genes across $u_s$ and $u_f$.
\textbf{Bottom:} Abagen‐derived expression maps of nine consensus genes plotted on Destrieux cortical (top rows) and subcortical (bottom rows) surfaces. Color scale denotes normalized expression (0–0.9). CHUK: Conserved Helix-Loop-Helix Ubiquitous Kinase, TH: Tyrosine Hydroxylase, TMEM106B: Transmembrane Protein 106B, MCL1: Myeloid Cell Leukemia 1, NOTCH1: Notch receptor 1}
\label{fig_gene}
\end{figure}

As shown in Fig. \ref{fig_gene}, the top six genes selected by frequency show a modest difference between SC- and FC-dominant patterns. Notably, the LASSO regression for FC-dominant patterns identified a consistent set of genes across both ADNI and OASIS datasets—including CHUK, NOTCH1, MCL1, HSPA5, and SP1 (Fig. \ref{fig_gene}, left, shaded). These genes were robustly selected across bootstrap resamples (blue sectors in the pie chart), suggesting a stable association with FC-dominant tau propagation. Many of the genes are involved in cellular stress responses, synaptic signaling, and neuroinflammation processes implicated in activity-dependent tau spread. In contrast, SC-dominant propagation was associated with distinct expression patterns of TH, MCL1, TMEM106B, NOTCH1, and IKBKB (Fig. \ref{fig_gene}, middle, shaded). These genes are enriched for functions supporting axonal structure and white matter integrity. For example, TMEM106B is involved in lysosomal trafficking and myelination; NOTCH1 and CTNNB1 regulate neurodevelopment and synaptic plasticity; IKBKB controls inflammatory cascades that may accelerate axonal degeneration; and HSPD1, a mitochondrial chaperonin, aids neuronal protein folding and protects against oxidative damage. Together, these findings indicate a biologically meaningful distinction: FC-dominant tau propagation reflects activity-driven vulnerability tied to synaptic and inflammatory pathways \cite{hetz2012unfolded,perciavalle2012anti}, while SC-dominant spread aligns with mechanisms maintaining axonal health and glial support. 

\textit{Discussion.} \textit{CHUK}, \textit{MCL1}, \textit{TH}, \textit{TMEM106B}, and \textit{NOTCH1} were shared across both SC- and FC-dominant gene lists, suggesting that these genes play a robust and substrate-independent role in tau propagation. Their consistent selection across models and datasets provides evidence that there exists a set of molecular drivers that may be fundamentally involved in the spread of tau pathology, regardless of whether it is shaped primarily by functional dynamics or structural pathways. This convergence supports the notion that both network types may act through common biological processes, such as inflammation (\textit{CHUK}, \textit{IKBKB}), mitochondrial function (\textit{MCL1}), neurodevelopment (\textit{NOTCH1}), and lysosomal trafficking (\textit{TMEM106B}), which together modulate vulnerability to tau dissemination across the brain.

Fig.~\ref{fig_gene} (bottom) visualizes the spatial expression gradients of these five consensus genes on Destrieux-parcellated cortical and subcortical surfaces, confirming that their expression patterns broadly co-localize with regions of high tau deposition. From the spatial maps, we observed that each gene exhibits a distinct but biologically plausible gradient that mirrors tau-vulnerable regions: 
\begin{itemize}
\item \textbf{\textit{CHUK}} expression exhibits striking regional specificity in the human brain, with relatively high levels observed in the medial prefrontal cortex, anterior cingulate cortex, and medial temporal lobe—including the hippocampus and entorhinal cortex—regions critically involved in cognitive control, emotional regulation, and memory processing. These areas are among the earliest affected by tau pathology and neurodegeneration in AD \cite{braak1991neuropathological, seeley2009neurodegenerative}. In contrast, CHUK expression is markedly lower in primary sensory and visual cortices, suggesting a preferential involvement in higher-order associative regions rather than unimodal processing areas. Notably, the expression pattern is bilaterally symmetric and includes deep cortical and limbic structures, supporting a potential role in global regulatory processes across hemispheres. Functionally, CHUK encodes IKK$\alpha$, a central kinase in the canonical NF-kB signaling pathway, which is a key mediator of inflammation and cellular stress responses. CHUK has been implicated in astrocyte and microglia activation under neuroinflammatory conditions \cite{ghosh2002missing}, processes known to contribute to region-specific vulnerability in AD \cite{heneka2015neuroinflammation}. Elevated CHUK expression in tau-vulnerable areas may thus reflect glial-driven innate immune activity that exacerbates tau pathology via cytokine release, oxidative stress, or blood–brain barrier dysfunction.
\item \textbf{\textit{TMEM106B}} exhibits moderate yet widespread cortical expression, with relative enrichment in the dorsolateral prefrontal cortex and angular gyrus—regions associated with executive and semantic functions. In subcortical areas, expression peaks in the hippocampal formation and limbic structures adjacent to the insula, which are known epicenters of early tau accumulation in Alzheimer’s disease \citep{braak1991neuropathological, maass2019alzheimer}. TMEM106B encodes a lysosomal membrane protein essential for endo‑lysosomal trafficking and autophagic clearance; its dysfunction exacerbates tau and TDP‑43 pathologies in cellular and animal models \citep{feng2024loss, jiao2023tmem106b}. Together, these spatial and mechanistic patterns suggest that TMEM106B may contribute to regional tau aggregation by impairing lysosomal.
\item \textbf{\textit{Tyrosine Hydroxylase (TH)}} shows highest expression in medial temporal regions—particularly the entorhinal cortex and parahippocampal gyrus—as well as in subcortical structures such as the hippocampus, amygdala, and striatum. Expression is comparatively lower in lateral parietal and frontal cortices. This spatial profile aligns with TH’s role in catecholamine biosynthesis and overlaps with regions susceptible to early tau pathology in Alzheimer's disease \cite{goralski2023spatial, braak1991neuropathological}.
\item \textbf{\textit{MCL1}} shows relatively low expression throughout the neocortex but exhibits pronounced upregulation in medial temporal and subcortical regions, particularly the hippocampus, amygdala, and thalamus. This spatial pattern is consistent with MCL1’s role in regulating mitochondrial-dependent apoptosis, a process implicated in tau-related neurodegeneration and neuronal survival in AD \cite{goralski2023spatial}.
\item \textbf{\textit{NOTCH1}} exhibits moderate expression in heteromodal association cortices, including the middle temporal and inferior parietal regions, as well as in subcortical limbic structures. This pattern aligns with NOTCH1’s established role in neural development and synaptic plasticity, processes that are increasingly recognized as relevant to neurodegenerative disease progression \cite{ables2011not,zhou2022}.

\end{itemize}

Furthermore, we identified the genetic risk factors that are linked to the SC-FC dominance selection pattern that both $u_s$ (tau propagation associated with SC) and $u_f$ (tau propagation associated with FC) exhibit, showing group differences between CN and AD. We first selected cortical regions that showed significant differences in $u_s$ and $u_f$ between AD and CN groups based on mixed-effects modeling. For these significant regions, we constructed a binary dominance index vector as the response variable: regions where SC predominated ($u_s > u_f$) were assigned a label of $+1$, while regions where FC predominated ($u_f > u_s$) were labeled $-1$. We then applied non-negative LASSO regression using this dominance vector as the target $y$ and the gene expression matrix as predictors, again recording the selection frequency $s_j$ of each gene across $B = 100$ bootstrap resamples. This approach allowed us to pinpoint AD risk genes whose spatial expression patterns are predictive of whether structural or functional pathways more strongly influence tau propagation in a given brain region. The top-right panel of Fig.~\ref{fig_gene} highlights risk genes identified through non-negative LASSO regression as predictive of SC- versus FC-dominant tau propagation. Interestingly, many of the top-selected genes overlapped with those identified in earlier $u_s$ and $u_f$–based analyses, suggesting that these genes not only track with tau burden but also differentiate the relative contribution of structural and functional pathways to tau spread. Among these, \textit{NOTCH1}, \textit{TH}, \textit{CHUK}, and \textit{MCL1} stood out as consistently selected across both analyses, reinforcing their relevance as core modulators of regional tau vulnerability. \textit{SP1}, a transcription factor known to regulate tau phosphorylation and amyloid precursor protein (APP) expression, was also robustly selected—pointing to its potential role at the intersection of amyloid and tau pathology. Several additional genes were identified that may refine the SC–FC dichotomy, including those involved in axonal transport (e.g., \textit{KIF5C}), immune signaling (e.g., \textit{TNFAIP3}), and calcium homeostasis (e.g., \textit{ATP2B2}). These findings collectively suggest that the structural–functional balance of tau propagation is shaped not only by macroscale connectome features but also by spatially distributed molecular programs that govern neuroinflammation, mitochondrial integrity, transcriptional regulation, and neuronal excitability.

\subsection{Mediation analysis for the interplay between SC-driven, FC-driven tau propagation, and cognitive decline in AD}
\label{sec_medi}
Since tau pathology directly reflects neurodegeneration and cognitive impairment in AD, Tau accumulation is associated with Mini-Mental State Examination (MMSE) scores \cite{jack2018nacc}. The disentangled tau accumulation $u_s$ driven by SC and $u_f$ by FC provides a new opportunity to elucidate the causal relationship between $u_s$, $u_f$, and MMSE. In this context, a set of structural equation models (SEMs) are listed below (Table \ref{tab:mediation_SEM2}):

\begin{table*}[h!]
\centering
\caption{Summary of mediation analysis paths and variables.}
\label{tab:mediation_SEM2}
\begin{tabular}{lcc}
\toprule
\textbf{Description} & \makecell[l]{Effect of $u_f$ on MMSE \\ mediated by $u_s$ (Fig. \ref{fig:mediation}a) }  & \makecell[l]{Effect of $u_s$ on MMSE \\ mediated by $u_f$ (Fig. \ref{fig:mediation}b) } \\
\midrule
Direct path     & $u_f \to MMSE$                     & $u_s \to MMSE$ \\
Indirect path   & $u_f \to u_s \to MMSE$             & $u_s \to u_f \to MMSE$ \\
Predictor $X$   & functional propagation $u_f$       & structural propagation $u_s$ \\
Mediator $M$    & structural propagation $u_s$       & functional propagation $u_f$ \\
Outcome $Y$     & MMSE                             & MMSE \\
Confounders     & age, sex, APOE4 status             & age, sex, APOE4 status \\
\bottomrule
\end{tabular}
\end{table*}

\begin{figure}[h!]
\centering
\includegraphics[width=1\textwidth, trim={0cm 0cm 0cm 0cm}, clip]{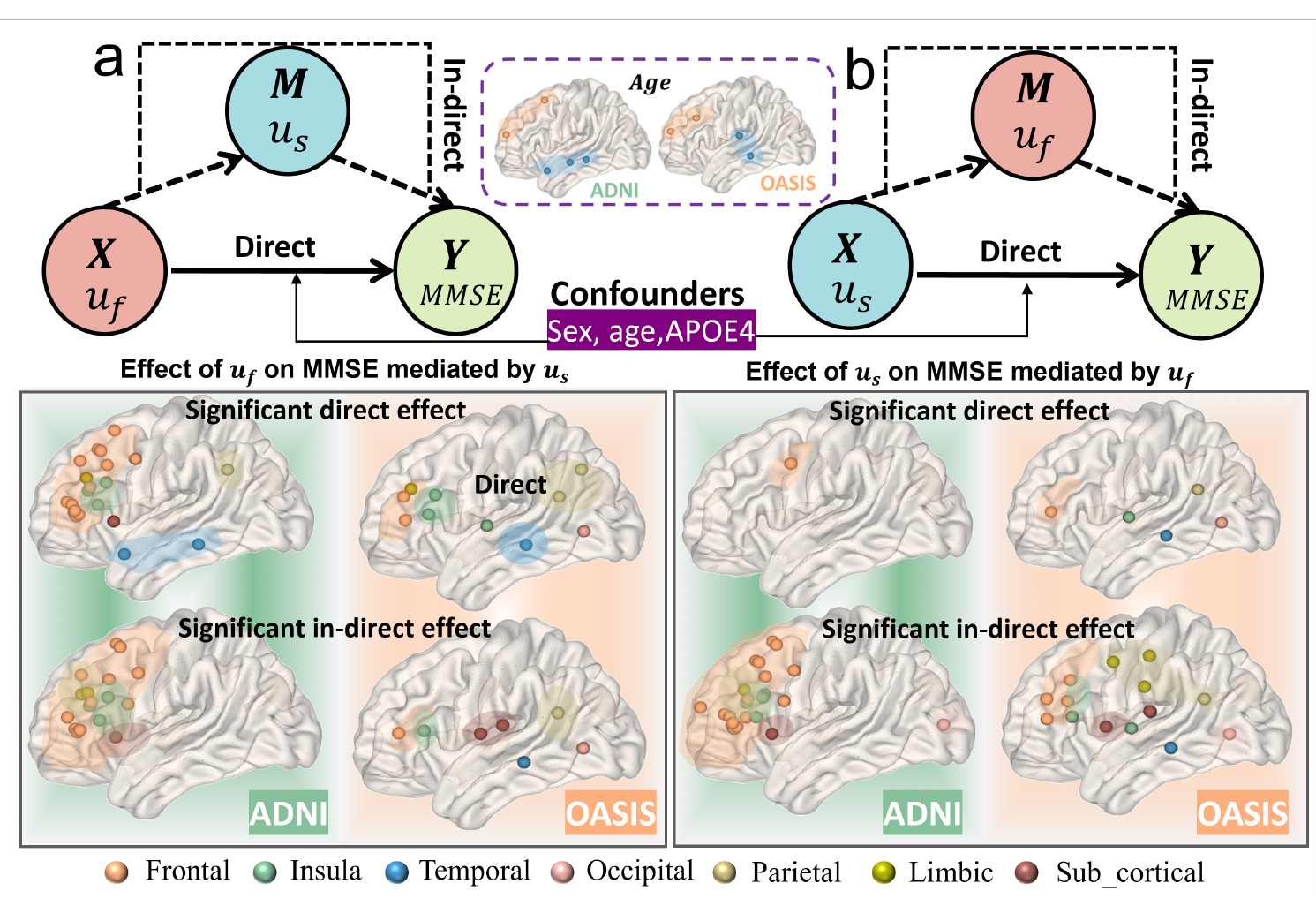}
\caption{\textbf{Mediation analysis of structural and functional contributions to cognition through tau propagation.} 
Top: SEM: $u_f\rightarrow u_s\rightarrow$MMSE (a) and SEM $u_s\rightarrow u_f\rightarrow$MMSE (b).
Bottom: Identified brain region showing significant direct effect ($1_{st}$ row) and in-direct effect ($2^{nd}$ row), at a significance level $p<0.05$, where we show the mediation result for ADNI (with green background) and OASIS (with red background) respectively. 
Age, sex, and APOE genotype are included as covariates. The purple dashed box highlights regions where age showed significant effects on MMSE ($p < 0.01$) in both datasets, particularly within the frontal and temporal lobes.
}
\label{fig:mediation}
\end{figure}

Brain regions exhibiting significant direct or indirect effects on MMSE at the $p<0.05$ level are displayed at the bottom of Fig. \ref{fig:mediation} (left: SEM results corresponding to the second column of Table \ref{tab:mediation_SEM2}; right: third column). Regions identified from the ADNI and OASIS datasets are highlighted with green and red backgrounds, respectively. 
While both SC and FC pathways contribute to tau accumulation, our findings indicate that functional connectivity plays a more prominent role in directly linking tau pathology to cognitive decline. In contrast, tau spread mediated by structural connectivity shows a substantially weaker association with cognitive impairment, indicating the critical influence of functional network disruption on cognitive outcomes.
In summary, neither SC nor FC alone fully explains cognition. Instead, global cognitive performance emerges from their dynamic interplay to the extent that white‐matter scaffolds seed functional co‐activations which in turn promote anatomical constraints to drive cognitive decline.

\textit{Discussion.} In addition to mediation analyses, we also examined the role of demographic and genetic confounders in the SEM. Among the covariates, age emerged as a significant factor associated with cognitive decline via tau propagation in the frontal and temporal lobes across both cohorts, pointing to increased susceptibility of these lobes to age-related cognitive decline. In contrast, APOE4 genotype and sex exhibited minimal regional effects, indicating a more modest modulatory role in the observed relationship between connectivity-driven tau propagation and cognitive decline. This pattern is consistent with prior studies demonstrating that age is a primary driver of cognitive decline in AD, particularly in frontal and temporal regions, while the effects of APOE4 and sex are often more indirect or context-dependent \cite{chen2009evaluation,sarin2018role,larsson2017type}.

\section{Method validation}

\subsection{Rigorous model validation across data cohorts}

To fully evaluate our model’s ability to forecast longitudinal tau burden, we trained and tested on baseline-to-follow-up SUVR data using a 5-fold cross-validation. We benchmarked against following leading deep-learning approaches on ADNI and OASIS datasets: (1) Graph neural networks including \textit{GCN} \cite{kipf2016semi}, \textit{GCNII} \cite{chen2020simple}, \textit{GAT}, and \textit{GATv2} \cite{brody2022how}.  For these methods, the baseline SUVR vector serves as node features, and SC and/or FC define the adjacency.
(2) Recurrent neural network (\textit{RNN}) \cite{nguyen2020predicting} (Sequential model), which ingests the baseline SUVR without explicit graph structure.  
(3) Liquid time-constant network (\textit{LTCNet}) \cite{hasani2021liquid} (Dynamical systems model), which is designed under a continuous‐time recurrent architecture.  

To disentangle the contributions of white‐matter versus functional pathways, we performed ablations using (i) SC alone, (ii) FC alone, and (iii) both SC and FC (“SC+FC”). For non-graph models, we concatenated separate SC- and FC-driven propagations via a final fully-connected layer. Table~\ref{tab:exper} reports mean absolute error (MAE) across all methods and input settings.  

Our multi‐layer neural transport model outperforms every competitor (paired $t$‐test, $p<0.05$) under all ablation conditions, demonstrating that explicitly modeling SC–FC interactions yields superior accuracy in predicting future tau burden.  

\begin{table*}[htbp]
\centering
\caption{Mean Absolute Error (MAE) of different models using SC, FC, or combined SC+FC inputs. Values are mean $\pm$ SD.}
\label{tab:exper}
\begin{small}
\begin{tabular}{llccc}
\toprule
\textbf{Dataset} & \textbf{Method} & \textbf{SC} & \textbf{FC} & \textbf{SC+FC} \\
\midrule
\multirow{7}{*}{ADNI}
 & GCN     & 0.204\,$\pm$\,0.047 & 0.193\,$\pm$\,0.026 & 0.189\,$\pm$\,0.011 \\
 & GCNII  & 0.221\,$\pm$\,0.021 & 0.315\,$\pm$\,0.039 & 0.347\,$\pm$\,0.040 \\
 & GAT     & 0.466\,$\pm$\,0.072 & 0.420\,$\pm$\,0.065 & 0.509\,$\pm$\,0.053 \\
 & GATv2   & 0.371\,$\pm$\,0.027 & 0.432\,$\pm$\,0.021 & 0.353\,$\pm$\,0.051 \\
 & RNN     & 0.334\,$\pm$\,0.051 & 0.368\,$\pm$\,0.029 & 0.291\,$\pm$\,0.046 \\
 & LTC‐Net & 0.238\,$\pm$\,0.032 & 0.252\,$\pm$\,0.067 & 0.234\,$\pm$\,0.029 \\
 & Our Multi-layer Model    & \bfseries0.093\,$\pm$\,0.023 & \bfseries0.134\,$\pm$\,0.036 & \bfseries0.062\,$\pm$\,0.014 \\
\midrule
 \multirow{7}{*}{OASIS3}
 & GCN     & 0.482\,$\pm$\,0.058 & 0.501\,$\pm$\,0.040 & 0.479\,$\pm$\,0.032 \\
 & GCNII  & 0.533\,$\pm$\,0.023 & 0.575\,$\pm$\,0.023 & 0.577\,$\pm$\,0.038 \\
 & GAT     & 0.526\,$\pm$\,0.027 & 0.570\,$\pm$\,0.016 & 0.517\,$\pm$\,0.029 \\
 & GATv2   & 0.471\,$\pm$\,0.033 & 0.551\,$\pm$\,0.024 & 0.495\,$\pm$\,0.026 \\
 & RNN     & 0.519\,$\pm$\,0.045 & 0.552\,$\pm$\,0.039 & 0.522\,$\pm$\,0.027 \\
 & LTC‐Net & 0.457\,$\pm$\,0.048 & 0.469\,$\pm$\,0.023 & 0.442\,$\pm$\,0.029 \\
 & Our Multi-layer Model    & \bfseries0.392\,$\pm$\,0.133 & \bfseries0.415\,$\pm$\,0.156 & \bfseries0.356\,$\pm$\,0.137 \\
\bottomrule
\end{tabular}
\end{small}
\end{table*}

\begin{figure}[h]
\centering
\includegraphics[width=1\textwidth]{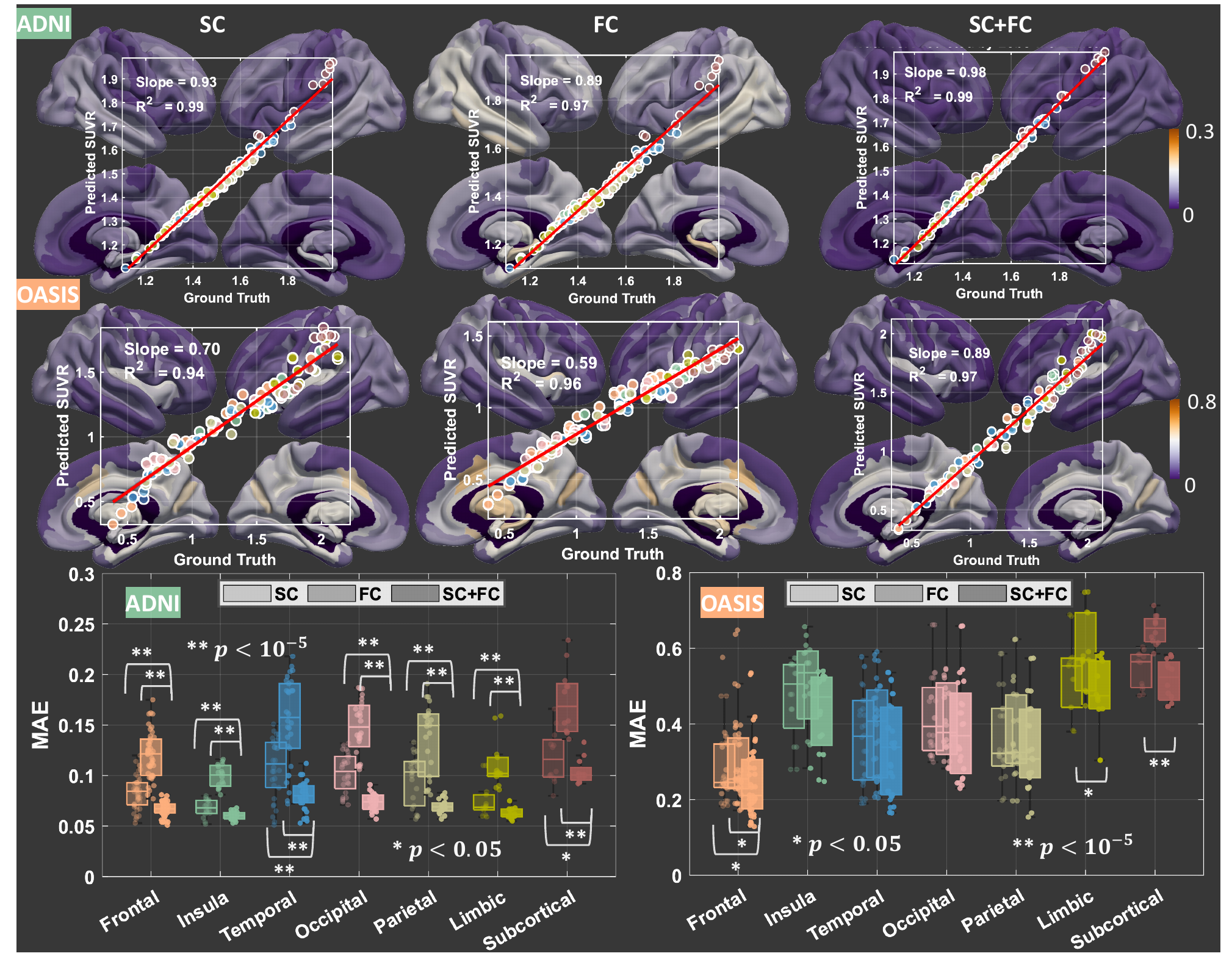}
\caption{\textbf{The predicted performance by our model on ADNI and OASIS datasets}. \textit{Top}: MAE between follow‐up PET‐derived tau distributions and model predictions based on SC alone (left), FC alone (middle), and the combined SC–FC interaction (right). Colors denote different cerebral lobes. The accompanying brain maps illustrate the regional average MAE for each modeling approach, with orange indicating higher error and purple indicating lower error. \textit{Bottom}: Lobar‐wise mean and standard deviation of MAE for predicting future tau burden using SC only, FC only, and SC–FC interaction.}\label{fig7}
\end{figure}

There is a clear gain in performance when both SC and FC are modeled jointly. To quantify their individual contributions, we performed an ablation study with our multi-layer transport model. Fig. \ref{fig7} (top) shows scatter plots of observed follow-up SUVR versus model predictions under three regimes—SC only (left), FC only (middle), and SC+FC interaction (right)—with each point colored by its anatomical lobe. Fitting a simple linear regression yields: SC only (ADNI: slope=0.93, $R^2=0.99$, OASIS: slope=0.70, $R^2=0.94$), FC only (ADNI: slope=0.89, $R^2=0.97$, OASIS: slope=0.59, $R^2=0.96$) and SC+FC (ADNI: slope=0.98, $R^2=0.99$, OASIS: slope=0.89, $R^2=0.97$). Our multi-layer model (SC+FC) not only recovers an unbiased (near unity) relationship to ground truth but also explains substantially more variance, indicating the critical role of SC–FC interactions in accurate tau forecasting. Fig. \ref{fig7} (bottom) summarizes regional prediction errors (MAE) averaged within each of the seven lobes (frontal, insula, temporal, occipital, parietal, limbic, subcortical). Boxplots denote MAE for SC only (left), FC only (middle), and SC+FC (right). Across every lobe—most notably temporal and subcortical regions—the combined SC–FC model consistently achieves the lowest error. Together, these ablation results provide converging evidence that explicitly modeling the interplay between anatomical wiring and functional co-activation is essential for maximizing the fidelity of tau‐propagation predictions.

\begin{figure}[h!]
\centering
\includegraphics[width=1\textwidth, trim={0cm 0cm 0cm 0cm}, clip]{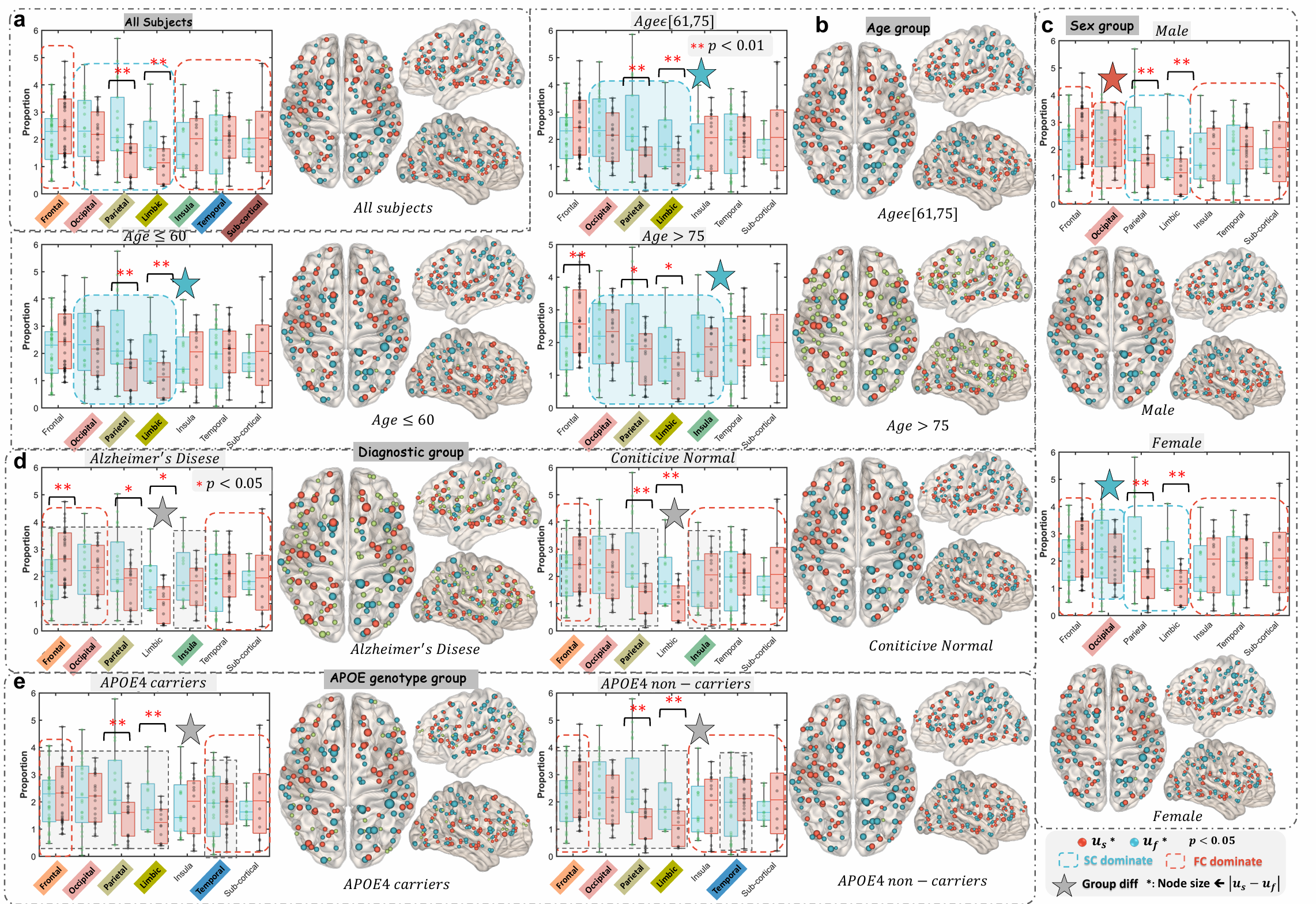}
\caption{\textbf{Group-wise analysis of structural vs. functional propagation on the OASIS dataset.}
\textbf{(a) All subjects.} \textit{Left}: Proportion of tau spread attributed to structural connectivity ($u_s^1$, blue) and functional connectivity ($u_f^1$, red) across brain lobes. \textit{Right}: Comparative map of $u_s^1$ versus $u_f^1$, where blue marks regions with stronger SC‐driven propagation and red marks those dominated by FC‐driven propagation. Node sizes scale with the absolute magnitude of the difference, $|u_s^1 - u_f^1|$.
\textbf{(b) Age groups.} SC and FC contributions to tau propagation across three age stages: $<$60, 61–75, and $>$75 years. With increasing age, tau propagation shifts toward greater reliance on structural connectivity, reflecting an age-dependent transition from FC- to SC-dominant pathways—a pattern that aligns with findings from the ADNI cohort.
\textbf{(c) Sex groups.} SC vs. FC propagation stratified by sex. Sex exerts only a minor modulatory effect on tau propagation, with a modest occipital difference that does not alter the overall SC-to-FC propagation architecture.
\textbf{(d) Diagnostic groups.} Group-wise propagation patterns across clinical diagnoses (e.g., CN, AD). A widespread shift in SC–FC dominance is observed between Alzheimer’s disease and cognitively normal groups, with most brain regions exhibiting either subtle or pronounced transitions—mirroring patterns reported in the ADNI cohort.
\textbf{(e) \textit{APOE4} status.} SC and FC contributions based on the presence or absence of the \textit{APOE4} allele. APOE-4 genotype modulates the balance between SC- and FC-mediated tau propagation, with carriers showing a regional shift toward SC dominance, consistent with ADNI findings.
}
\label{oasisscfc}
\end{figure}

Although anatomical tracts are known being associated with the prion-like tau propagation theory, our multilayer model shows that functional co‑activation can further amplify accumulation—particularly in later‑affected regions (temporal lobe and subcortical nuclei), which are FC‑dominant. This suggests a stage-dependent shift from predominantly SC-driven seeding to FC-mediated propagation in advanced AD. To validate the generalizability of the identified tau propagation patterns, we further examined SC versus FC contributions using the independent OASIS dataset (Fig. \ref{oasisscfc}). Consistent with our main analyses, tau propagation showed differential reliance on SC and FC networks across cortical regions, with a notable transition from FC-dominant to SC-dominant pathways with increasing age. Stratified analyses by sex revealed only subtle modulatory effects, with minimal sex-related variations largely restricted to occipital regions. In contrast, comparisons across clinical diagnostic groups demonstrated significant shifts toward structural dominance in AD relative to cognitively normal subjects, aligning closely with patterns observed in the ADNI cohort. Additionally, carriers of the APOE4 allele exhibited greater reliance on SC-mediated tau propagation, underscoring genetic modulation of SC–FC interactions. These convergent findings from OASIS strongly support the robustness and consistency of our observed tau propagation mechanisms across diverse populations.

\section{Conclusion and future work}
In this study, we developed a novel multi-layer neural transport model to investigate how structural and functional connectivity jointly shape the propagation of pathological tau in Alzheimer's disease. By explicitly modeling SC–FC interactions, our framework captures complex, region-specific dynamics of tau spread underlining the topology of human connectome. Our findings reveal that functional connectivity primarily drives tau propagation in subcortical, insular, frontal, and temporal regions, whereas structural connectivity plays a greater role in occipital, parietal, and limbic areas. Overall, the relative contribution of structural and functional connectivity to tau propagation evolves throughout the course of disease progression, where FC predominates in the early stages while SC plays a more prominent role as the disease advances. Moreover, we demonstrate that the spatial distribution of SC- and FC-dominant regions aligns with the expression patterns of AD risk genes, including CHUK, TMEM106B, MCL1, NOTCH1, and TH. Meanwhile, we observe dynamic shifts in SC–FC contributions across disease stages, regionally specific modulation by APOE genotype and A$\beta$ deposition, and subtle sex- and age-related alterations in the dominant pathways of tau propagation. These findings are robust across independent AD cohorts and offer mechanistic insights into the network-level basis of disease progression. 

Our current computational framework is based on the assumption that tau propagation is shaped by the underlying network topology. Although we explore the influence of other biomarkers, such as A$\beta$ plaques, on tau propagation in post-hoc analyses, additional AD-relevant biomarkers have not yet been incorporated into the model.
In future work, we plan to generalize our modeling framework to take biomarker-to-biomarker interactions into account. We will apply our multi-layer model to other neurodegenerative disorders to uncover shared or distinct principles of pathological spread across brain networks.

\section{Methods}
\label{sec:methods}

\subsection{Problem Definition}
We aim to characterize how large-scale brain network architecture shapes the propagation of tau pathology throughout the course of Alzheimer’s disease (AD). Each subject’s structural and functional connectomes are represented as a weighted graph $\mathcal{G} = (V, W)$, where $V = \{v_i\}_{i=1}^N$ denotes $N$ brain regions and $W = [w_{ij}]_{i,j=1}^N$ encodes connection strengths. Regional tau burden is quantified using standardized uptake value ratios (SUVRs), denoted as $x = [x_i]_{i=1}^N$. Given baseline tau $x^0$, our goal is to predict future tau $x^1$ by leveraging both anatomical and functional connectivity. The structural connectivity (SC) matrix $S = [s_{ij}]$ and functional connectivity (FC) matrix $F = [f_{ij}]$ serve as the graph adjacency representations.

\subsection{Tau Propagation as Network-Guided Transport}
Neuropathological evidence suggests that tau does not diffuse randomly, but preferentially spreads along large-scale brain networks. To mechanistically capture this phenomenon, we embed tau burden into an energy-based latent space $u = \phi(x)$, where $\phi$ is a learnable mapping. The spatiotemporal evolution of $u$ follows a conservative mass-transport principle, analogous to diffusion:
\begin{equation}
\frac{\partial u}{\partial t} + \mathrm{div}(q) = 0,
\end{equation}
where $\mathrm{div}(\cdot)$ denotes the graph divergence and $q$ represents the flux along edges. Following Fourier’s law:
\begin{equation}
q = -c \nabla u, \quad (\nabla u)_{ij} = w_{ij}(u_i - u_j),
\end{equation}
where $c$ denotes diffusivity and $\nabla$ is the graph gradient. Substituting into the continuity equation gives:
\begin{equation}
\frac{\partial u}{\partial t} = -\mathrm{div}(c \nabla u) = -c \Delta u, \quad u=\phi(x),
\end{equation}
where $\Delta$ is the graph Laplacian. This formulation provides a physics-inspired and biologically informed model of how connectome topology influences tau spread.

\begin{figure}[!t]
    \centering
    \includegraphics[width=\textwidth]{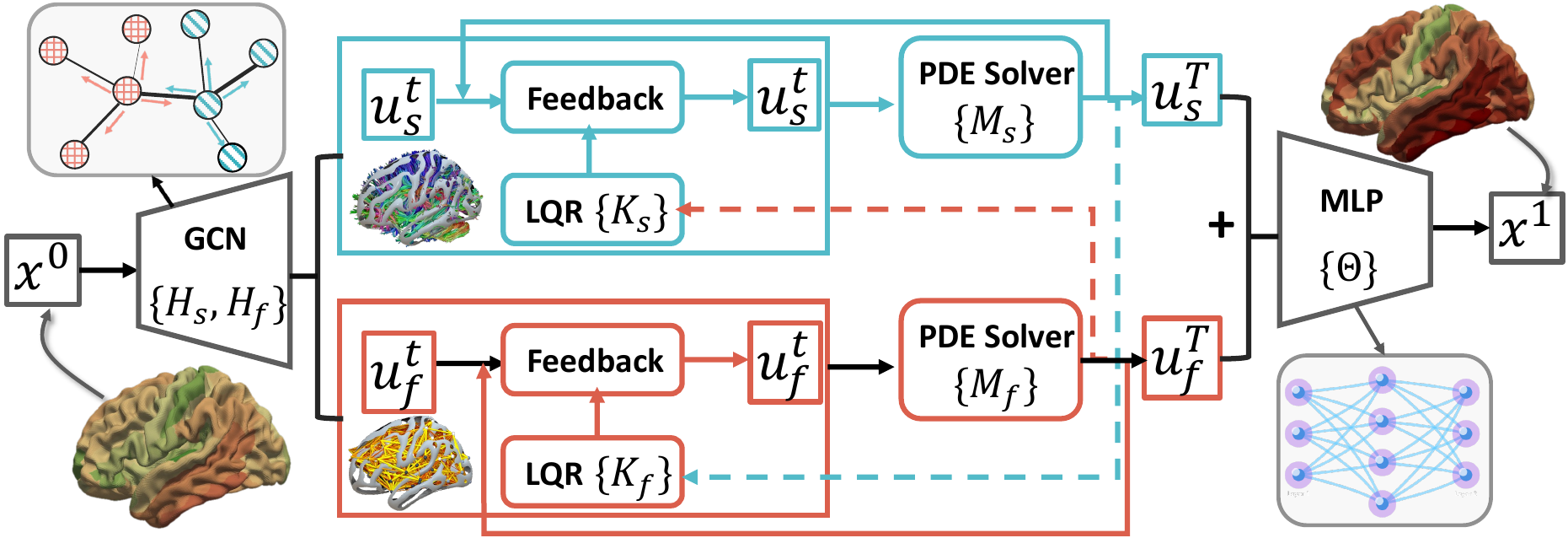}
    \caption{\textbf{Architecture of the proposed closed-loop multi-layer neural transport model.} Baseline tau SUVR values $x^0$ are mapped to SC- and FC-specific latent potentials $u_s, u_f$ governed by coupled transport equations with feedback control. The final output $\hat{x}^1$ predicts tau at the next time point. The design enables mechanistic interpretability of SC vs.\ FC propagation pathways.}
    \label{fig:multilayer_model}
\end{figure}
\subsection{Multi-Layer SC–FC Propagation Model}
Tau pathology may propagate along distinct but interacting anatomical and functional pathways. To account for this, we introduce two coupled latent states, $u_s$ and $u_f$, representing SC-driven and FC-driven propagation, respectively. Each state evolves on its own connectivity network while interacting through learnable cross-layer terms, forming a bi-layer network model (Fig.~\ref{fig:multilayer_model}). Let $\Delta_s$ and $\Delta_f$ denote the Laplacians of the SC and FC graphs:
\begin{equation}
\frac{\partial u_s}{\partial t} = -c\,\Delta_s u_s, \qquad 
\frac{\partial u_f}{\partial t} = -c\,\Delta_f u_f.
\end{equation}

To model cross-talk between layers, we introduce bidirectional feedback via trainable coupling matrices $M_s, M_f$ and weights $\lambda_s, \lambda_f$:
\begin{equation}
\begin{cases}
\frac{\partial u_s}{\partial t} = -\Delta_s u_s + \lambda_s M_s u_f, \\
\frac{\partial u_f}{\partial t} = -\Delta_f u_f + \lambda_f M_f u_s.
\end{cases}
\end{equation}
This can be expressed compactly as:
\begin{equation}
\frac{\partial}{\partial t}
\begin{bmatrix} u_s \\ u_f \end{bmatrix} =
\underbrace{\begin{bmatrix}-\Delta_s & \lambda_s M_s \\ \lambda_f M_f & -\Delta_f\end{bmatrix}}_{\mathcal{A}}
\begin{bmatrix} u_s \\ u_f \end{bmatrix},
\end{equation}
with the constraint $\phi^{-1}(u_s + u_f) = x$.

\subsection{Closed-Loop Control for SC--FC Coupling}
We incorporate a feedback control formulation to dynamically balance the contributions of SC and FC over time. This mechanism learns subject- and region-specific weights, indicating when and where tau propagation is predominantly anatomical versus functional. Designed for mechanistic interpretability, the latent variables $u_s$ and $u_f$ explicitly capture SC- and FC-driven propagation components, allowing the detection of regional shifts in dominant pathways as the disease progresses. Such shifts can be quantitatively examined against known biological and genetic factors, including APOE genotype, amyloid burden, and inflammation-related gene expression. Formally, we cast the coupled dynamics as a controllable linear system and apply a full-state feedback control framework:
\begin{equation}
\frac{\partial u}{\partial t} = \Delta u - M K u, \quad M=[M_s,M_f],
\end{equation}
where feedback gains $K_s, K_f$ regulate SC- and FC-driven propagation. The energy cost functional is:
\begin{equation}
\mathcal{L} = \tfrac{1}{2} u_s^\top Q u_s + \tfrac{1}{2} u_f^\top R u_f,
\end{equation}
where $Q, R$ are diagonal weight matrices. Using Pontryagin’s Minimum Principle, we define:
\begin{equation}
\mathcal{J}(u) = \tfrac{1}{2} u^\top P u, \quad 
P=\begin{bmatrix}P_{ss} & P_{sf}\\ P_{sf}^\top & P_{ff}\end{bmatrix}.
\end{equation}
The optimal feedback law has closed-form $K=-R^{-1} M P$, where $P$ satisfies the algebraic Riccati equation:
\begin{equation}
P \mathcal{A} + \mathcal{A}^\top P + \begin{bmatrix}Q & 0\\0 & R\end{bmatrix}=0.
\end{equation}
This decomposition allows separate optimization of intra-layer ($P_{ss}, P_{ff}$) and cross-layer ($P_{sf}$) effects.

\subsection{Learning Strategy and Implementation}
The learnable parameters include:
$\Gamma=\{H_s,H_f,M_s,M_f,K_s,K_f\},$
where $H_s,H_f$ map SUVR to potential energy, $M_s,M_f$ capture SC--FC feedback, and $K_s,K_f$ encode feedback control. The coupled PDE is numerically integrated to obtain $u_s,u_f$, and a supervised neural network maps $(u_s,u_f)$ to predicted tau $\hat{x}^1$. The objective minimizes:
$\|x^1 - \hat{x}^1\|_F^2,$ ensuring faithful prediction while maintaining biological interpretability of SC--FC contributions.

\section{Resource Availability}
\textbf{Data Availability.} The involved disease-based data can be found and downloaded in the Image and Data Archive (IDA) -\href{https://adni.loni.usc.edu/}{ADNI}. OASIS3 can be found in \href{https://sites.wustl.edu/oasisbrains/home/oasis-3/}{here}.

\section{Author contributions}
Conceptualization, T.D. and G.W.; methodology,  T.D., and G.W.; 
software, X.H. and Y.Z.; 
validation, T.D. and Y.Z.; 
formal analysis, T.D., X.H. and J.D.; investigation, T.D. and X.H.; 
data curation, T.D. X.H., J.D.and Y.Z.; 
writing—original draft, T.D., X.H. G.W. and J.D.; writing—review and editing, T.D., X.H. and G.W.; visualization, T.D. and X.H.; 
supervision, G.W.;
project administration, G.W.

\section{Declaration of interests}
The authors declare no competing interests.

\begin{appendices}

\section{Experimental results}
\label{supp_results}

\begin{figure}[h!]
\centering
\includegraphics[width=0.98\textwidth, trim={0cm 0cm 0cm 0cm}, clip]{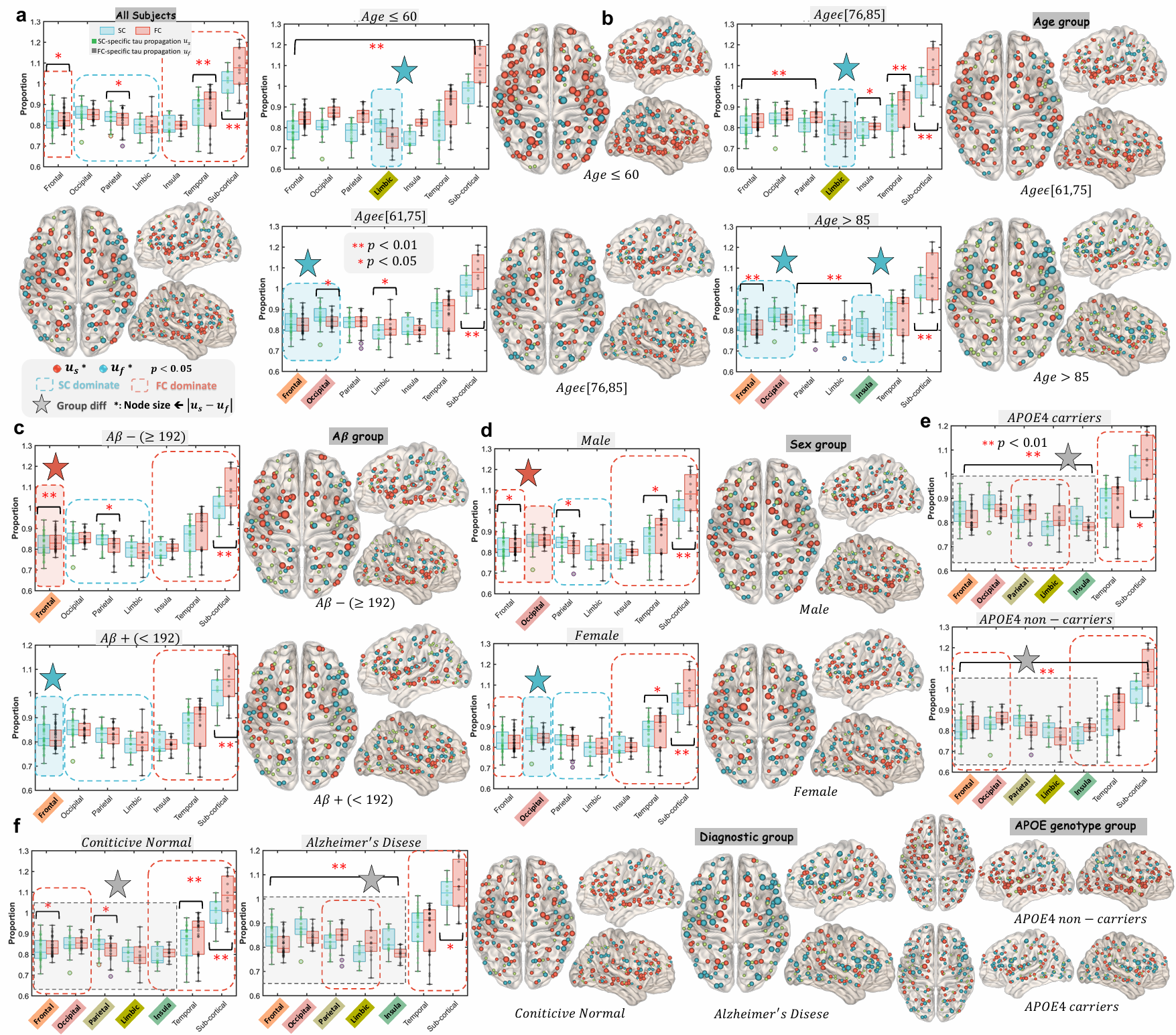}
\caption{\textbf{Group-wise analysis of structural vs. functional propagation on ADNI dataset.} \textbf{(a) all subjects.} \textit{Left}: Fraction of tau propagation attributed to SC ($u_s$, blue) and FC ($u_f$, red) across brain lobes.  
\textit{Right}: Comparison between $u_s$ and $u_f$. Blue nodes indicate regions where SC-driven tau propagation ($u_s$) exceeds FC-driven propagation ($u_f$), while red nodes indicate the opposite. Node size reflects the magnitude of the difference, $|u_s - u_f|$.
 \textbf{(b) Age groups:} SC and FC contributions to tau propagation across four age stages: $<$60, 61–75, 76–85, and $>$85 years. Age-dependent shift in tau propagation, with younger individuals showing FC-dominant spread and older individuals showing increasing reliance on SC pathways.
\textbf{(c) Amyloid-$\beta$ status:} Comparisons between $A\beta+$ (defined as $A\beta < 192$) and $A\beta-$ ($A\beta \ge 192$) individuals. A$\beta$ burden “boosts” tau spread along functional circuits (especially in frontal cortex). 
\textbf{(d) Sex groups:} SC vs. FC propagation stratified by sex. Sex exerts only a minor modulatory effect on tau propagation, with a modest occipital difference that does not alter the overall SC-to-FC propagation architecture.
\textbf{(e) \textit{APOE4} status:} SC and FC contributions based on presence or absence of the \textit{APOE4} allele. APOE4 carriers show a transition from FC- to SC-dominant tau spread in the frontal, occipital and insula cortex.
\textbf{(f) Diagnostic groups:} Group-wise propagation patterns across clinical diagnoses (e.g., CN, AD). Except for the temporal and subcortical regions, all other lobes exhibited a reversal in SC–FC dominance between APOE-4/AD and non-carrier/CN groups.
}\label{adniscfc3}
\end{figure}

We present detailed statistical analyses in Fig. \ref{adniscfc3} based on the proposed multi-layer neural transport model, quantifying the region-specific and stage-dependent contributions of structural and functional connectivity to tau propagation.

\section{Data proprecessing}\label{secA1}
\subsection{Neuroimage data preprocessing}
\label{data-pre}

\paragraph{Data Description}
We evaluate our proposed method using two publicly available datasets: the Alzheimer's Disease Neuroimaging Initiative (ADNI) \cite{ADNI} and the Open Access Series of Imaging Studies-3 (OASIS3) \cite{marcus2007open}, both widely used in Alzheimer's disease research.

\textbf{ADNI dataset.}  
The ADNI dataset includes 539 tau-PET scans and 1012 processed structural connectomes. For our study, we selected 163 subjects who have diffusion-weighted imaging (DWI), fMRI, and longitudinal tau-PET scans (2--5 time points), ensuring that each subject has complete tau, structural, and functional connectivity data.

\textbf{OASIS3 dataset.}  
OASIS3 comprises 1379 subjects with 2842 MRI sessions and 2157 PET sessions. From this dataset, we identified 81 subjects with DWI, fMRI, and longitudinal tau-PET scans (2 time points) to provide a comparable multi-modal dataset for model evaluation.




Table \ref{Demographic} lists the detailed demographic statistics for ADNI and OASIS3 (highlighted in gray shadow) data. We show age, MMSE (mini-mental state examination for baseline), gender, and longitudinal PET image acquisition interval period (from baseline to the last follow-up visit). 

\begin{table*}[htbp]
\caption{Demographic statistics for ADNI and OASIS3 (shaded) data. M/F denotes male/female and Y denotes year.}
\label{Demographic}
\setlength{\tabcolsep}{3pt}
\begin{center}
\begin{small}
\begin{tabular}{lcccc|c}
\toprule
\textbf{ADNI} & \textbf{Age} & \textbf{MMSE}  & \textbf{Gender} & \textbf{Period} & \textbf{APOE4} \\
\midrule
Mean$\pm$std& 75.1$\pm$7.7 &  28.0$\pm$2.0  & 48\%/52\%  & 1.5$\pm$0.5 & 23,33,43/24,34,44   \\ 
Range& 55 $\sim$ 94 &  20 $\sim$ 30  & 261/278 (M/F)  & 1$\sim$4 (Y) & 68.4\%/31.6\%\\ 
\hline
\hline
\rowcolor{gray!20}\textbf{OASIS3} & \textbf{Age} & \textbf{MMSE}  & \textbf{Gender}  & \textbf{Period} & \textbf{APOE4}\\
 \hline
\rowcolor{gray!20}Mean$\pm$std & 61.8$\pm$7.1 & 29.0$\pm$1.0  & 71.6\%/28.4\%  & 0.6$\pm$0.5 & 23,33,43/24,34,44   \\ 
\rowcolor{gray!20} Range& 46 $\sim$ 77 &  24 $\sim$ 30  & 58/23 (M/F) & 0$\sim$1 (Y)  & 62.9\%/37.1\%\\ 

\bottomrule
\end{tabular}
\end{small}
\end{center}
\end{table*}

\subsection*{Multimodal Imaging Processing Pipelines}
\begin{figure}[h!]
\centering
\includegraphics[width=0.9\textwidth]{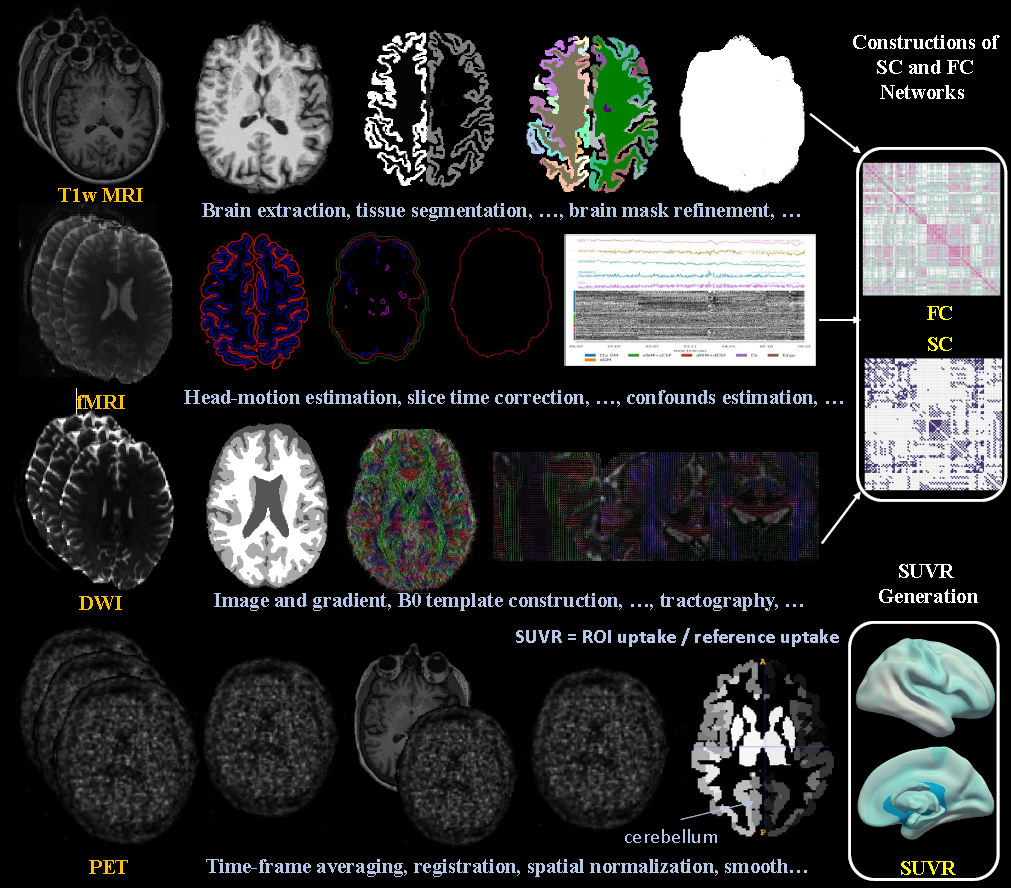}
\caption{Workflow of generating FC/SC matrices and SUVR values.}\label{Pre}
\end{figure}

The data preprocessing (Fig. \ref{Pre}) involves the following steps to derive FC/SC matrices and SUVR values:

$\bullet$ \textbf{Functional Connectivity (FC) construction from fMRI\footnote{\url{https://fmriprep.org/en/stable/}}:}

$\triangleright$ \textit{Structural MRI (T1-weighted) preprocessing}: includes brain extraction, tissue segmentation, spatial normalization, cost function masking, longitudinal processing, and brain mask refinement.  

$\triangleright$ \textit{BOLD preprocessing}: comprises reference image creation, head-motion correction, slice-timing correction, susceptibility distortion correction, EPI-to-T1 registration, resampling to standard spaces, projection onto FreeSurfer surfaces and HCP Grayordinates, and confound estimation.  

$\triangleright$ \textit{BOLD time-series extraction}: used for constructing functional connectivity matrices.

    
        
         

$\bullet$ \textbf{SC Construction from DWI\footnote{\url{https://qsiprep.readthedocs.io/en/latest/}}:}
    
$\triangleright$ \textit{Initial processing}, including image and gradient orientation conformity, distortion grouping, denoising, distortion correction, head motion correction, and B0 template construction.
        
$\triangleright$ \textit{Reconstruction steps}, such as ODF/FOD estimation, anisotropy scalar computation, and tractography.

$\bullet$ \textbf{SUVR Generation from PET:}

$\triangleright$ \textit{Frame selection and averaging:} From the 4D PET series, select tracer‐specific equilibrium frames (e.g.\ 80–100 min post‐injection) and average them into a single static image.

$\triangleright$  \textit{Motion correction:} Rigid‐body register frames (or the static average) to correct head motion.

$\triangleright$ \textit{MRI co‐registration:} Align the motion‐corrected PET to the subject’s T1w MRI via mutual‐information registration.

 $\triangleright$ \textit{Spatial normalization:} Warp the co‐registered PET into a standard template (e.g.\ MNI152).
 
 $\triangleright$ \textit{Smoothing:} Apply a Gaussian kernel (e.g.\ 6–8 mm full width at half maximum (FWHM)).
 
$\triangleright$ \textit{SUVR computation:} Divide each ROI uptake by the mean uptake in a reference region (e.g.\ cerebellar gray matter).

$\triangleright$ \textit{Feature extraction:} Use an anatomical (Destrieux \cite{destrieux2010automatic}) to extract mean SUVRs per region or network.

$\triangleright$  \textit{Quality control and analysis:} Inspect registration and SUVR maps; exclude poor‐quality scans; perform statistical tests (group‐ or network‐based).

\subsection{Gene Expression Data Processing} 
\label{gene_exp}

\begin{figure}[h!]
\centering
\includegraphics[width=1\textwidth]{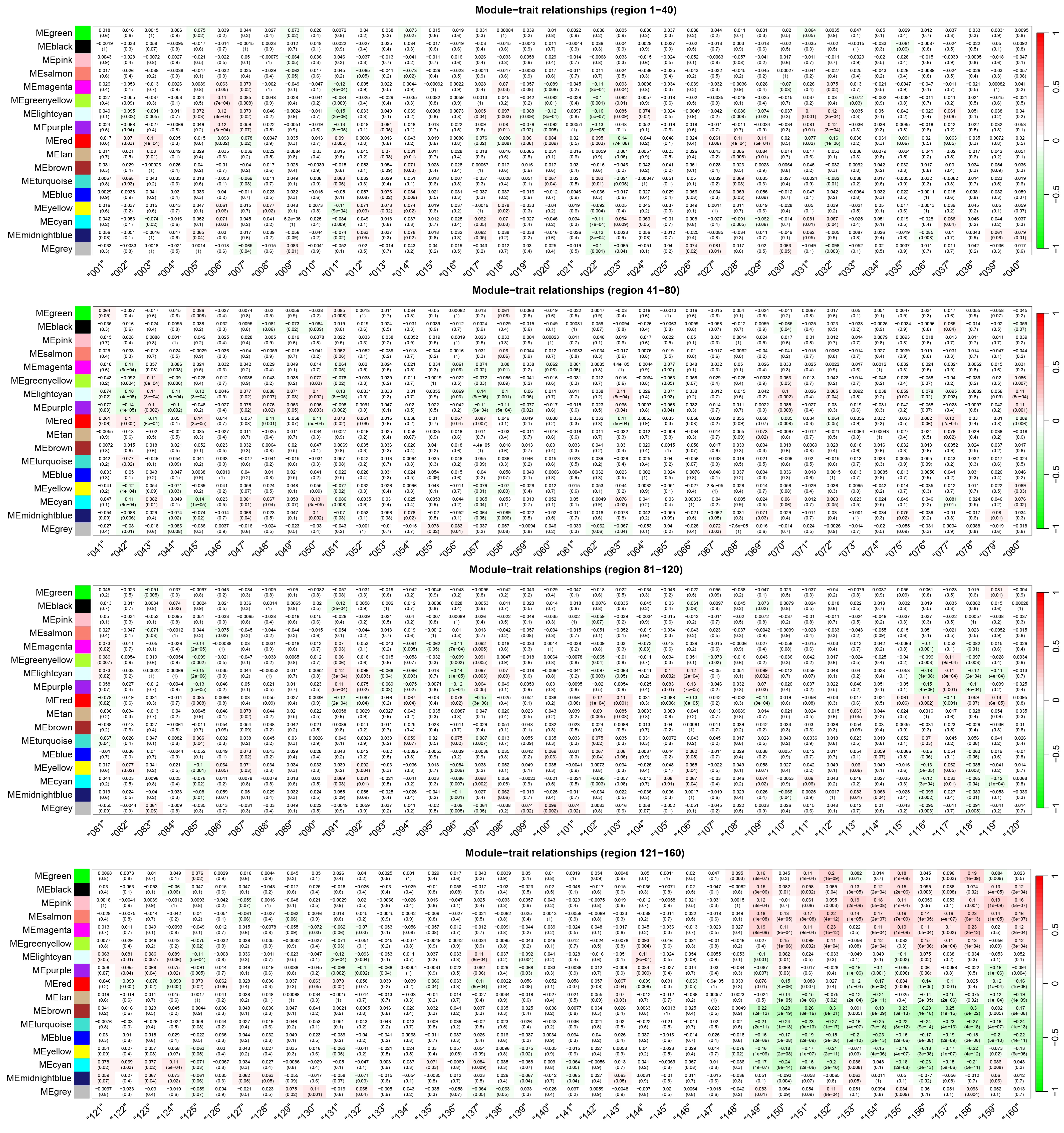}
\caption{\textbf{Identification of modules associated with the 160 brain regions.} Heatmap of the correlation between the module eigengenes and brain regions. We selected the ME salmon-grade block for subsequent analysis. Red means over-expression, green indicates under-expression; numbers in each cell give the corresponding $t$-test $p$-value. The numbers on the axis (001-160) correspond to the brain regions in Table \ref{label_name}.}\label{gene_1}
\end{figure}
Regional microarray expression data were obtained from six post-mortem adult human brains (one female; age range: 24.0--57.0 years; mean age: $42.50 \pm 13.38$) provided by the Allen Human Brain Atlas (AHBA) \cite{hawrylycz2012anatomically}. Data processing was conducted using the \href{https://github.com/rmarkello/abagen}{\texttt{abagen} toolbox} (v0.1.3) with a 160-region volumetric brain atlas in MNI space. This atlas comprises 148 cortical regions from the volumetric Destrieux atlas \cite{destrieux2010automatic} and 12 subcortical regions from the volumetric Desikan-Killiany atlas \cite{DESIKAN2006968}.

\paragraph{Probe Selection}
Microarray probes were reannotated using updated mappings provided by \cite{arnatkeviciute2019practical}, and any probes without valid Entrez Gene IDs were discarded. Subsequently, probes were filtered based on their expression intensity relative to background noise \cite{quackenbush2002microarray}; probes with intensity below background in more than 50\% of samples were removed, resulting in 31,569 retained probes.

For genes represented by multiple probes, the probe with the most consistent spatial expression pattern across donors was selected based on differential stability \cite{hawrylycz2015canonical}:
\begin{equation} 
\Delta_S(p) = \frac{1}{\binom{N}{2}} \sum_{i=1}^{N-1} \sum_{j=i+1}^{N} \rho[B_i(p), B_j(p)],
\end{equation}
where $\rho$ denotes Spearman's rank correlation of probe $p$ expression across regions in donors $B_i$ and $B_j$, and $N$ is the total number of donors.

\paragraph{Region Assignment}
To assign tissue samples to brain regions:
\begin{itemize}
\item \textbf{Coordinate update:} Tissue sample MNI coordinates were updated using non-linear registration via \href{https://github.com/chrisfilo/alleninf}{ANTs} 
\item \textbf{Sample augmentation:} Samples were mirrored bilaterally across hemispheres to improve spatial coverage \cite{romero2018structural}.
\item \textbf{Tolerance:} Samples were assigned to a region if their MNI coordinate was within 2 mm of a parcel.
\item \textbf{Misassignment prevention:} Sample-to-region assignment was restricted by hemisphere and gross anatomical division (e.g., cortex, subcortex/brainstem, cerebellum) \cite{arnatkeviciute2019practical}.
\item \textbf{Missing regions:} Regions without any assigned sample were filled by mapping each voxel in the region to its nearest donor sample. Regional expression was estimated as a distance-weighted average over all voxels.
\item \textbf{Unassigned samples:} Samples not assigned to any atlas-defined region were discarded.
\end{itemize}

\paragraph{Normalization}
To address inter-subject variability, tissue sample expression values were normalized across genes using a robust sigmoid function \cite{fulcher2013highly}:
\begin{equation}
x_{\text{norm}} = \frac{1}{1 + \exp\left(-\frac{(x - \langle x \rangle)}{\text{IQR}_x}\right)}
\end{equation}
where $\langle x \rangle$ is the sample median and $\text{IQR}_x$ is the interquartile range. These normalized values were rescaled to the unit interval:
\begin{equation}
x_{\text{scaled}} = \frac{x_{\text{norm}} - \min(x_{\text{norm}})}{\max(x_{\text{norm}}) - \min(x_{\text{norm}})}
\end{equation}

The same procedure was applied across tissue samples for each gene. For each donor, gene expression values were averaged across samples assigned to the same brain region, yielding a 160-region by 15,633-gene expression matrix.

\paragraph{Gene expression data selection}
\begin{figure}[h!]
\centering
\iflightpdf
\includegraphics[width=0.72\textwidth]{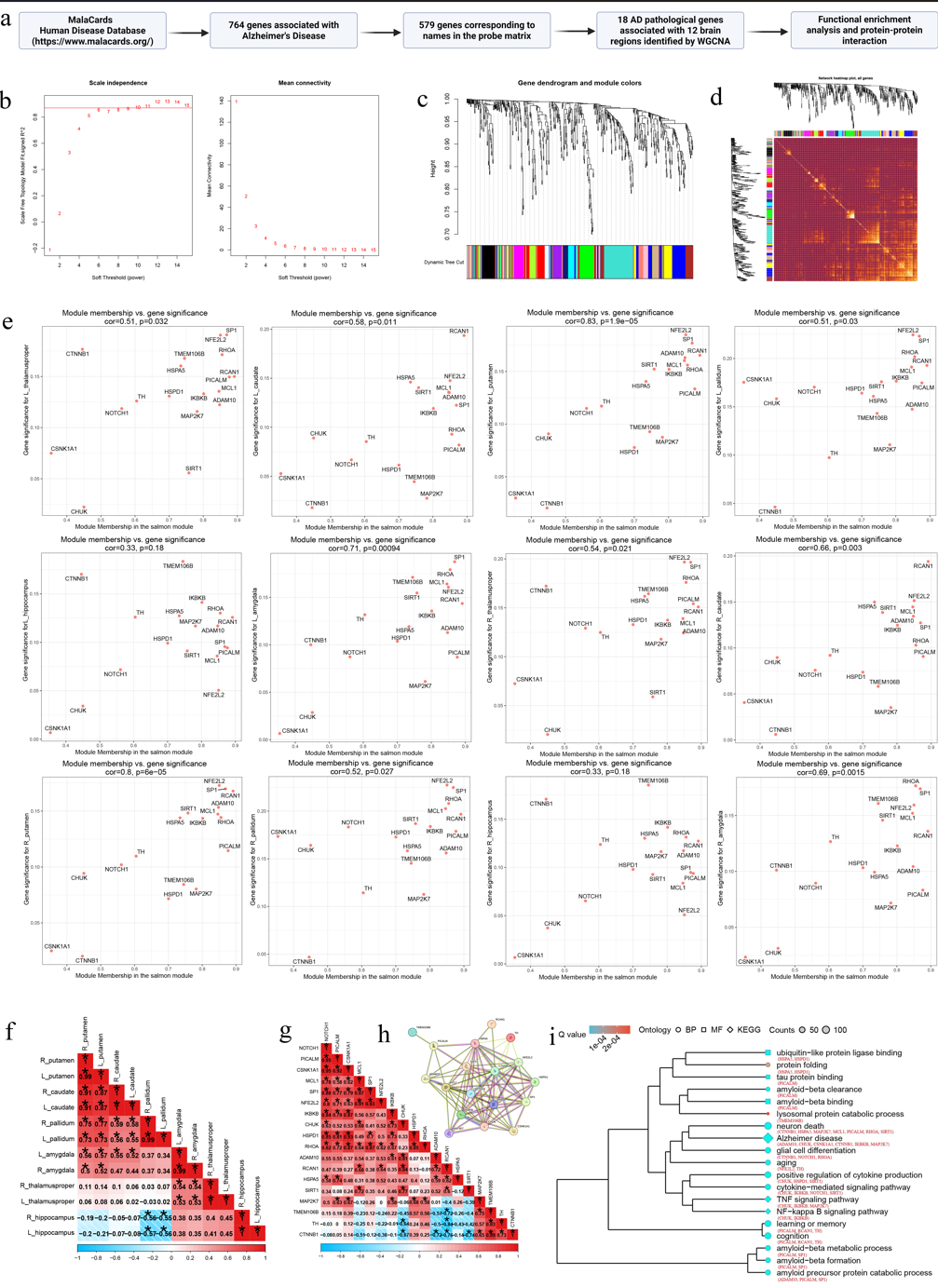}
\caption{\textbf{Identification of the important Alzheimer's disease genes among 160 brain regions}. (a) Screening process for Alzheimer's disease pathology genes. (b) Choosing the best soft-threshold power. This analysis includes the scale-free index for various soft-threshold powers ($\beta$) and the mean connectivity for various soft-threshold powers. (c) Hierarchical clustering dendrogram of genes for determining consensus modules. The color band exhibits the results obtained from the automatic single-block analysis. (d) Heatmap plot of topological overlap in the gene network. In this heatmap, each row and column represents a gene, with lighter colors indicating low topological overlap and progressively darker red shades indicating higher topological overlap. The darker squares along the diagonal correspond to modules. The gene dendrogram and module assignments are displayed on the left and top sides, respectively. (e) A scatterplot illustrating the relationship between gene significance (GS) for the corresponding brain region and module membership (MM) in the salmon module. The analysis reveals a highly significant correlation between GS and MM, indicating that hub genes within the salmon module are also likely to have a strong association with the corresponding brain region. (f) Spearman correlation analysis between 12 brain regions based on the expression matrices of 18 Alzheimer's disease genes. (g) Spearman correlation analysis between 18 Alzheimer's disease genes based on their expression matrices across 12 brain regions. (h) Protein-Protein interaction analysis for 18 Alzheimer's disease genes. (i) The pathological functions of 18 Alzheimer's disease genes. $*p<$0.05.}\label{gen0}
\end{figure}

\begin{figure}[h!]
\centering
\iflightpdf
\includegraphics[width=1\textwidth]{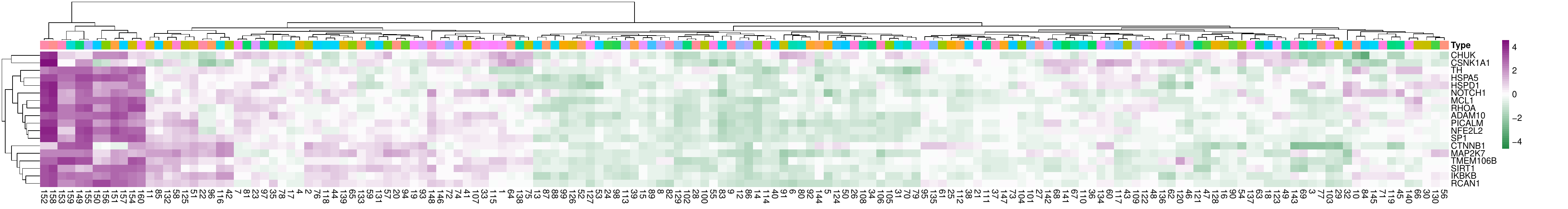}
\caption{\textbf{Heatmap of average expression value of 18 Alzheimer's disease genes across 160 Abagen-derived brain regions.}}\label{gene_2}
\end{figure}

We first selected 579 Alzheimer’s disease (AD)–associated genes from an initial pool of 15,633 genes, based on their spatial correlation with tau pathology. We then performed weighted gene co-expression network analysis (WGCNA) and functional enrichment assessment on these genes using the Abagen-derived gene expression atlas mapped to 160 Destrieux parcels (Table \ref{label_name}). As illuminated in Fig. \ref{gen0}a, we obtained 764 recognized genes associated with AD from MalaCards Human Disease Database (https://www.malacards.org/). A total of 579 genes corresponding to the names in the Abagen-derived gene expression probe matrix were utilized for WGCNA. In this analysis, the soft-threshold power of this gene co-expression network construction was calibrated to 9 (scale-free $R^2 = 0.90$) (Fig. \ref{gen0}b). Sixteen meaningful modules were identified, except that the gray module is a set of genes that are not suitable for any specified module (Fig. \ref{gen0}c, d). Among the 16 modules, we found that the salmon module, which contains 18 AD-related genes including RCAN1, NOTCH1, PICALM, ADAM10, HSPD1, NFE2L2, HSPA5, MCL1, MAP2K7, CHUK, IKBKB, CTNNB1, RHOA, SIRT1, SP1, TH, TMEM106B, and CSNK1A1, shows a positive association with 12 brain regions, including thalamusproper (149, left; 155, right), caudate (150, left; 156, right), putamen (151, left; 157, right), pallidum (152, left; 158, right), hippocampus (153, left; 159, right), and amygdala (154, left; 160, right) (Fig. \ref{gene_2}). All of these genes demonstrated a positive relationship with the module and most of the 12 brain regions (Fig. \ref{gen0}e). In comparison to the other 148 brain regions, these 18 genes exhibited consistently high and specific expression levels in the 12 brain regions (Fig. \ref{gene_1}). Spearman correlation analysis conducted on the expression matrices of 18 genes across 12 brain regions revealed a highly positive correlation between the expression profiles of corresponding left and right brain regions ($R > 0.99$, $p < 0.05$), suggesting no significant difference in the expression of the 18 genes between the left and right brain regions (Fig. \ref{gen0}f). In addition, most of the 18 genes were observed to exhibit significant correlations at the mRNA level (Fig. \ref{gen0}g). Protein-Protein interaction analysis at the protein level also confirmed the existence of interactions among those genes (Fig. \ref{gen0}h). Furthermore, functional enrichment assessment exhibited that those genes are involved in aging, cognition, learning or memory (PICALM, RCAN, and TH), AD, and multiple AD-related pathological processes such as tau protein binding (PICALM), protein folding (HSPA5 and HSPD1), ubiquitin-like protein ligase binding (HSPA5 and HSPD1), lysosomal protein catabolic process (TMEM106B), amyloid precursor protein catabolic process (ADAM10, PICALM, and SP1), neuron death (CTNNB1, HSPA5, MAP2K7, MCL1, PICALM, RHOA, and SIRT1), and etc. (Fig. \ref{gen0}i). Those findings indicate that the above brain regions and AD genes may play the crucial roles in the onset of AD. 

\begin{table}[htbp]
\centering
\tiny
\caption{Destrieux parcellation labels with region IDs (001–160).}
\label{label_name}
\setlength{\tabcolsep}{1pt} 
\begin{tabular}{|c|l|c|l|c|l|c|l|}
\hline
\textbf{ID} & \textbf{Label} & \textbf{ID} & \textbf{Label} & \textbf{ID} & \textbf{Label} & \textbf{ID} & \textbf{Label} \\
\hline
001 & L G\_and\_S\_frontomargin & 002 & L G\_and\_S\_occipital\_inf & 003 & L G\_and\_S\_paracentral & 004 & L G\_and\_S\_subcentral \\
005 & L G\_and\_S\_transv\_frontopol & 006 & L G\_and\_S\_cingul-Ant & 007 & L G\_and\_S\_cingul-Mid-Ant & 008 & L G\_and\_S\_cingul-Mid-Post \\
009 & L G\_cingul-Post-dorsal & 010 & L G\_cingul-Post-ventral & 011 & L G\_cuneus & 012 & L G\_front\_inf-Opercular \\
013 & L G\_front\_inf-Orbital & 014 & L G\_front\_inf-Triangul & 015 & L G\_front\_middle & 016 & L G\_front\_sup \\
017 & L G\_Ins\_lg\_and\_S\_cent\_ins & 018 & L G\_insular\_short & 019 & L G\_occipital\_middle & 020 & L G\_occipital\_sup \\
021 & L G\_oc-temp\_lat-fusifor & 022 & L G\_oc-temp\_med-Lingual & 023 & L G\_oc-temp\_med-Parahip & 024 & L G\_orbital \\
025 & L G\_pariet\_inf-Angular & 026 & L G\_pariet\_inf-Supramar & 027 & L G\_parietal\_sup & 028 & L G\_postcentral \\
029 & L G\_precentral & 030 & L G\_precuneus & 031 & L G\_rectus & 032 & L G\_subcallosal \\
033 & L G\_temp\_sup-G\_T\_transv & 034 & L G\_temp\_sup-Lateral & 035 & L G\_temp\_sup-Plan\_polar & 036 & L G\_temp\_sup-Plan\_tempo \\
037 & L G\_temporal\_inf & 038 & L G\_temporal\_middle & 039 & L Lat\_Fis-ant-Horizont & 040 & L Lat\_Fis-ant-Vertical \\
041 & L Lat\_Fis-post & 042 & L Pole\_occipital & 043 & L Pole\_temporal & 044 & L S\_calcarine \\
045 & L S\_central & 046 & L S\_cingul-Marginalis & 047 & L S\_circular\_insula\_ant & 048 & L S\_circular\_insula\_inf \\
049 & L S\_circular\_insula\_sup & 050 & L S\_collat\_transv\_ant & 051 & L S\_collat\_transv\_post & 052 & L S\_front\_inf \\
053 & L S\_front\_middle & 054 & L S\_front\_sup & 055 & L S\_interm\_prim-Jensen & 056 & L S\_intrapariet\_and\_P\_trans \\
057 & L S\_oc\_middle\_and\_Lunatus & 058 & L S\_oc\_sup\_and\_transversal & 059 & L S\_occipital\_ant & 060 & L S\_oc-temp\_lat \\
061 & L S\_oc-temp\_med\_and\_Lingual & 062 & L S\_orbital\_lateral & 063 & L S\_orbital\_med-olfact & 064 & L S\_orbital-H\_Shaped \\
065 & L S\_parieto\_occipital & 066 & L S\_pericallosal & 067 & L S\_postcentral & 068 & L S\_precentral-inf-part \\
069 & L S\_precentral-sup-part & 070 & L S\_suborbital & 071 & L S\_subparietal & 072 & L S\_temporal\_inf \\
073 & L S\_temporal\_sup & 074 & L S\_temporal\_transverse & 075 & R G\_and\_S\_frontomargin & 076 & R G\_and\_S\_occipital\_inf \\
077 & R G\_and\_S\_paracentral & 078 & R G\_and\_S\_subcentral & 079 & R G\_and\_S\_transv\_frontopol & 080 & R G\_and\_S\_cingul-Ant \\
081 & R G\_and\_S\_cingul-Mid-Ant & 082 & R G\_and\_S\_cingul-Mid-Post & 083 & R G\_cingul-Post-dorsal & 084 & R G\_cingul-Post-ventral \\
085 & R G\_cuneus & 086 & R G\_front\_inf-Opercular & 087 & R G\_front\_inf-Orbital & 088 & R G\_front\_inf-Triangul \\
089 & R G\_front\_middle & 090 & R G\_front\_sup & 091 & R G\_Ins\_lg\_and\_S\_cent\_ins & 092 & R G\_insular\_short \\
093 & R G\_occipital\_middle & 094 & R G\_occipital\_sup & 095 & R G\_oc-temp\_lat-fusifor & 096 & R G\_oc-temp\_med-Lingual \\
097 & R G\_oc-temp\_med-Parahip & 098 & R G\_orbital & 099 & R G\_pariet\_inf-Angular & 100 & R G\_pariet\_inf-Supramar \\
101 & R G\_parietal\_sup & 102 & R G\_postcentral & 103 & R G\_precentral & 104 & R G\_precuneus \\
105 & R G\_rectus & 106 & R G\_subcallosal & 107 & R G\_temp\_sup-G\_T\_transv & 108 & R G\_temp\_sup-Lateral \\
109 & R G\_temp\_sup-Plan\_polar & 110 & R G\_temp\_sup-Plan\_tempo & 111 & R G\_temporal\_inf & 112 & R G\_temporal\_middle \\
113 & R Lat\_Fis-ant-Horizont & 114 & R Lat\_Fis-ant-Vertical & 115 & R Lat\_Fis-post & 116 & R Pole\_occipital \\
117 & R Pole\_temporal & 118 & R S\_calcarine & 119 & R S\_central & 120 & R S\_cingul-Marginalis \\
121 & R S\_circular\_insula\_ant & 122 & R S\_circular\_insula\_inf & 123 & R S\_circular\_insula\_sup & 124 & R S\_collat\_transv\_ant \\
125 & R S\_collat\_transv\_post & 126 & R S\_front\_inf & 127 & R S\_front\_middle & 128 & R S\_front\_sup \\
129 & R S\_interm\_prim-Jensen & 130 & R S\_intrapariet\_and\_P\_trans & 131 & R S\_oc\_middle\_and\_Lunatus & 132 & R S\_oc\_sup\_and\_transversal \\
133 & R S\_occipital\_ant & 134 & R S\_oc-temp\_lat & 135 & R S\_oc-temp\_med\_and\_Lingual & 136 & R S\_orbital\_lateral \\
137 & R S\_orbital\_med-olfact & 138 & R S\_orbital-H\_Shaped & 139 & R S\_parieto\_occipital & 140 & R S\_pericallosal \\
141 & R S\_postcentral & 142 & R S\_precentral-inf-part & 143 & R S\_precentral-sup-part & 144 & R S\_suborbital \\
145 & R S\_subparietal & 146 & R S\_temporal\_inf & 147 & R S\_temporal\_sup & 148 & R S\_temporal\_transverse \\
149 & L\_thalamusproper & 150 & L\_caudate & 151 & L\_putamen & 152 & L\_pallidum \\
153 & L\_hippocampus & 154 & L\_amygdala & 155 & R\_thalamusproper & 156 & R\_caudate \\
157 & R\_putamen & 158 & R\_pallidum & 159 & R\_hippocampus & 160 & R\_amygdala \\
\hline
\end{tabular}

\end{table}

Specific analysis methods are as follows:

\begin{itemize}
    \item \textbf{Extraction of gene expression atlas for hub Alzheimer's disease genes.}  
    A total of 764 recognized genes associated with AD were obtained from the MalaCards Human Disease Database (https://www.malacards.org/), which aggregates data from multiple sources to present detailed profiles of various diseases. By matching these genes with those in the Abagen-derived gene matrix, the expression matrix of 579 genes was extracted for subsequent analysis.

    \item \textbf{GO and KEGG pathway enrichment analysis of 579 genes.}  
    Gene Ontology (GO) and KEGG enrichment analyses were conducted using the \texttt{clusterProfiler v4.4.4} R package to explore the biological functions and pathways associated with the 579 AD-related genes. Terms with a $p < 0.05$ were considered significantly enriched.

    \item \textbf{Weighted Gene Co-expression Network Analysis (WGCNA).}  
    WGCNA was performed on the expression matrix of the 579 AD-related genes using the WGCNA R package \cite{langfelder2008wgcna}. Genes and samples were filtered using the \texttt{goodSamplesGenes} function. A scale-free network was constructed with a soft threshold of $\beta = 9$ (achieving a scale-free fit $R^2 = 0.9$). The adjacency matrix was transformed into a topological overlap matrix (TOM), and hierarchical clustering (minimum module size = 10) was applied to define gene modules. Module eigengenes were calculated to summarize expression profiles per module and relate them to 160 brain regions.

    \item \textbf{Protein-protein interaction analysis.}  
    PPI analysis for 18 AD-related genes was conducted using the STRING database (https://string-db.org/) with the default confidence score threshold. The resulting networks were visualized using Cytoscape v3.10.2.

    \item \textbf{Spearman correlation analysis.}  
    Associations between 18 AD genes and 12 brain regions—including left/right thalamusproper (149, 155), caudate (150, 156), putamen (151, 157), pallidum (152, 158), hippocampus (153, 159), and amygdala (154, 160)—were evaluated using Spearman correlation via the R package \texttt{ggplot2 v3.4.4}. Associations with $p < 0.05$ were considered statistically significant.
\end{itemize}

\end{appendices}

\bibliography{sn-bibliography}

\end{document}